\documentclass[11pt, a4paper, logo, copyright]{dmml}

\pdftrailerid{redacted}

\usepackage[authoryear, sort&compress, round]{natbib}

\usepackage{times}
\usepackage{latexsym}

\usepackage[T1]{fontenc}

\usepackage[utf8]{inputenc}

\usepackage{microtype}

\usepackage{fix-cm}  

\usepackage{graphicx}
\usepackage[most]{tcolorbox}
\usepackage{url}
\usepackage{amsmath, amssymb}
\usepackage{multirow} 
\usepackage[table]{xcolor}
\usepackage{booktabs}
\usepackage{pifont}

\usepackage{enumitem}
\usepackage{ulem}
\usepackage{amsmath,amssymb,amsthm}
\usepackage{pifont}
\usepackage{graphicx}
\usepackage{pifont}
\usepackage{subcaption}
\usepackage{url}
\usepackage{xcolor}
\usepackage[table]{xcolor} 

\usepackage[utf8]{inputenc} 
\usepackage[T1]{fontenc}    
\usepackage[colorlinks=true]{hyperref}
\hypersetup{
    linkcolor=purple,       
    citecolor=purple,      
    filecolor=purple,    
    urlcolor=purple         
}
\usepackage{url}            
\usepackage{booktabs}       
\usepackage{amsfonts}       
\usepackage{nicefrac}       
\usepackage{microtype}      
\usepackage{xcolor,colortbl}         
\usepackage{enumitem}
\usepackage[nameinlink,capitalize,noabbrev]{cleveref}
\usepackage{titletoc}

\definecolor{lightlav}{HTML}{EDE7F6}
\definecolor{darklav}{HTML}{5E35B1}
\usepackage{graphicx}
\usepackage{booktabs}
\usepackage{siunitx}
\usepackage[table]{xcolor}
\usepackage{enumitem}
\usepackage{pifont}
\usepackage{amsthm}

\theoremstyle{definition}
\newtheorem{judgedef}{Definition}[section]
\usepackage{wrapfig}
\usepackage{caption}
\usepackage[table]{xcolor}
\definecolor{headercolor}{RGB}{219,217,226}

\usepackage{hyperref}      
\usepackage{fontawesome5}  
\usepackage{academicons}   
\usepackage{xcolor}        
\newcommand{\hf}{{\includegraphics[height=1.0em]{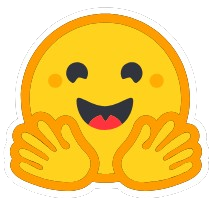}}}

\title{
  \raisebox{-0.7em}{\includegraphics[height=2em]{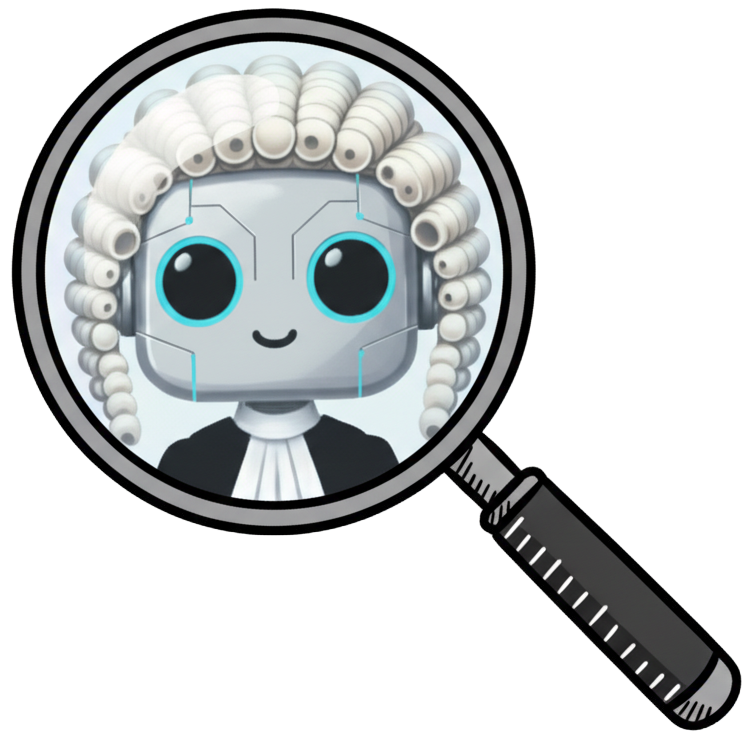}}
  Who’s Your Judge? On the Detectability of LLM-Generated Judgments
}

\correspondingauthor{daweili5@asu.edu}

\renewcommand{\today}{}

\author[1]{Dawei Li}
\author[1]{Zhen Tan}
\author[1]{Chengshuai Zhao}
\author[1]{Bohan Jiang}
\author[2]{Baixiang Huang}
\author[1]{Pingchuan Ma}
\author[1]{Abdullah Alnaibari}
\author[2]{Kai Shu}
\author[1]{Huan Liu}

\affil[1]{Arizona State University}
\affil[2]{Emory University}

\begin{abstract}
Large Language Model (LLM)-based judgments leverage powerful LLMs to efficiently evaluate candidate content and provide judgment scores. However, the inherent biases and vulnerabilities of LLM-generated judgments raise concerns, underscoring the urgent need for distinguishing them in sensitive scenarios like academic peer reviewing. In this work, we propose and formalize the task of judgment detection and systematically investigate the detectability of LLM-generated judgments. Unlike LLM-generated text detection, judgment detection relies solely on judgment scores and candidates, reflecting real-world scenarios where textual feedback is often unavailable in the detection process. Our preliminary analysis shows that existing LLM-generated text detection methods perform poorly given their incapability to capture the interaction between judgment scores and candidate content---an aspect crucial for effective judgment detection. Inspired by this, we introduce \textit{J-Detector}, a lightweight and transparent neural detector augmented with explicitly extracted linguistic and LLM-enhanced features to link LLM judges' biases with candidates' properties for accurate detection. Experiments across diverse datasets demonstrate the effectiveness of \textit{J-Detector} and show how its interpretability enables quantifying biases in LLM judges. Finally, we analyze key factors affecting the detectability of LLM-generated judgments and validate the practical utility of judgment detection in real-world scenarios.
\end{abstract}

\begin{document}
\maketitle

\vspace{0.5em}
\noindent\faGithub\ \href{https://github.com/David-Li0406/Judgment-Detection}{https://github.com/David-Li0406/Judgment-Detection}\\[0.3em]
\hf\ \href{https://huggingface.co/datasets/wjldw/JD-Bench}{https://huggingface.co/datasets/wjldw/JD-Bench}\\[0.3em]
\faGlobe\ \href{https://llm-as-a-judge.github.io}{https://llm-as-a-judge.github.io/}

\section{Introduction}
Taking advantage of the powerful Large Language Models (LLMs), the paradigm of LLM-based judgment~\citep{zheng2023judging,li2024generation} has been proposed, designed to automate and scale up various annotation and reviewing applications~\citep{lee2023rlaif,zhu2025deepreview,chang2025treereview}. By combining powerful LLMs with well-designed prompting strategies, LLM-based judgment enables human-like evaluation of long-form and open-ended generation in a more cost-efficient manner. For example, LLM-based judgment has been increasingly used in the peer review of leading AI conferences~\citep{liang2024monitoring}.


Despite this remarkable progress, many recent studies point out various biases of LLM-generated judgment toward spurious features, such as length and affinity~\citep{ye2024justice,li2025preference,zhao2025chain}. Besides, the vulnerability of the LLM judgment system has also been revealed, that several maliciously-designed and hard-to-detect tokens or words can fool the LLM judges to give much inconsistent scores despite the candidates' genuine quality~\citep{shi2024optimization,zhao2025one}. Recently, in the scenario of academic peer reviewing, some researchers sneak prompts, which are usually concealed as white text on a white background, into their papers to instruct LLMs to only provide positive feedback and thus trick AI reviewers\footnote {\url{https://www.theregister.com/2025/07/07/scholars\_try\_to\_fool\_llm\_reviewers/}}. 
All these challenges highlight the importance of distinguishing LLM-generated judgments to guarantee the assessment's fairness and reliability.

\begin{wrapfigure}{!th}{0.60\textwidth}
    \vspace{-6mm}
    \centering
    \includegraphics[width=\linewidth]{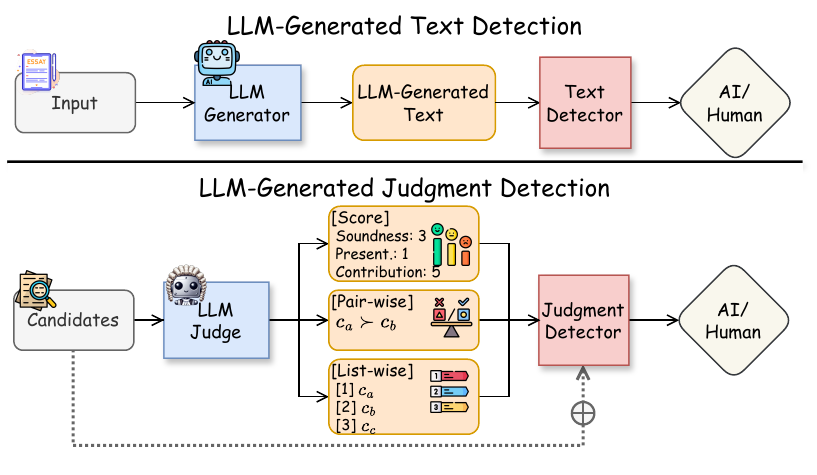}
    \vspace{-6mm}
    \caption{Comparison between LLM-generated judgment detection and text detection.}
    \label{fig:illustration}
    \vspace{-4mm}
\end{wrapfigure}
To address this concern, we propose the judgment detection task, which aims at examining the detectability of LLM-generated judgments across diverse scenarios. Unlike existing machine-generated text detection task that focuses on textual content~\citep{mitchell2023detectgpt}, judgment detection targets at distinguishing LLM-generated from human-produced judgments solely based on the \textit{candidate content} and \textit{judgment scores} (as illustrated in \cref{fig:illustration}). For instance, in academic paper reviewing, judgment detection will be performed using only the candidate paper and its assigned ratings (e.g., soundness, novelty, overall score), without accessing the full review text. This setting is particularly important for real-world scenarios where textual feedback is often unavailable in the detection process. For example, reviewers who adopt AI-generated reviews may intentionally submit minimal textual content, such as ``N/A'' to evade detection. Moreover, in the evaluation data labeling scenario, annotators are typically required to provide only the judgment scores. Score-based judgment detection is especially critical in these scenarios to identify the illegal use of LLM-generated judgment and guarantee assessment reliability.

Developing a good LLM-generated judgment detector is not trivial. In our warm-up analysis, we identify two key types of information for judgment detection which are not jointly considered in existing related approaches: \ding{182} \textbf{Judgment-Intrinsic Features}, which capture patterns within the judgment score distribution, and \ding{183} \textbf{Judgment-Candidate Interaction Features}, which capture the interaction between judgment scores and candidate content. Building on them, we find that existing LLM-generated text detection methods fail to capture Judgment-Candidate Interaction Features, leading to subpar performance---especially in single-dimension settings, where each judgment consists of a single score assessing one aspect of the candidates. To address this, we introduce \textit{J-Detector}, a lightweight and interpretable neural detector designed specifically for LLM-generated judgment detection. \textit{J-Detector} is augmented with explicitly extracted linguistic and LLM-enhanced features to capture systematic correlations between judgment scores and candidate features that LLM judges are often biased toward, thereby effectively leveraging these biases for more accurate detection.

Experiments across diverse judgment datasets demonstrate the effectiveness of \textit{J-Detector} and the two types of augmented features. Besides, we showcase how to leverage the interpretability of \textit{J-Detector} to enable bias quantification in LLM judges. Finally, we analyze key factors affecting the detectability of LLM-generated judgments and demonstrate a real-world application that integrates judgment detection with text-based detection to identify AI-generated reviews in an academic peer reviewing scenario.
In summary, our key contributions are:
\begin{itemize}[leftmargin=0pt, itemsep=0pt, topsep=0pt, parsep=0pt, partopsep=0pt]
\item We propose, for the first time, the judgment detection task, which aims at distinguishing human and LLM judgments based on judgment scores and candidate content.
\item We design \textit{J-Detector}, a lightweight and interpretable detection method, that effectively bridges candidate and judgment information with linguistic and LLM-enhanced features.
\item Through extensive experiments, we demonstrate the advantages of \textit{J-Detector}, identify key factors driving judgment detectability, and show the utility of judgment detection in real-world applications.
\end{itemize}

\section{Related Work}

\noindent\textbf{LLM-as-a-judge}, first introduced by \citet{zheng2023judging}, leverages powerful LLMs~\cite{zhang2024shifcon,zhang2024balancing,wang2024bpo} to automatically evaluate candidate content and assign scores as judgment results. This paradigm has been expanded to diverse applications to judge various types of candidates, including paper quality assessing~\citep{jin2024agentreview}, document relevance measurement~\citep{gao2023human,rahmani2024llmjudge}, and reasoning trace correctness verification~\citep{zhanggenerative}, driving substantial progress in automatic assessment~\citep{li2025smoa,tan2025prospect,beigi2024model,hueditable,hu2024understanding,jeong2024bluetempnet}. Despite these advances, recent studies highlight notable limitations. Research has uncovered systematic biases in LLM-generated judgments, where evaluations are influenced by spurious features such as response length or superficial affinity rather than genuine content quality~\citep{ye2024justice,li2025preference,jiang2024assessing,yang2025quantifying}. Moreover, adversarial work demonstrates that LLM judges can be manipulated with a few carefully crafted, hard-to-detect tokens or phrases, which induce disproportionately high scores misaligned with actual candidate quality~\citep{shi2024optimization,zhao2025one}. To mitigate these issues, methods such as bias quantification~\citep{ye2024justice} and human-in-the-loop calibration~\citep{wang2023large} have been proposed. Building on this line of research, we introduce a new task, judgment detection, that aims to distinguish and prevent the misuse of LLM-generated judgments.


\noindent\textbf{AI-generated Text Detection}
aims to distinguish machine-generated from human-produced text, evolving from early stylometric and perplexity-based methods~\citep{gehrmann2019gltr, zellers2019grover} to supervised classifiers~\citep{ippolito2020automatic, mitchell2023detectgpt}, and more recently toward robust, generalizable approaches such as zero-shot prompting and watermarking~\citep{sun2025zero,mao2025watermarking}.
Another relevant line of work for us is the detection of LLM-generated peer reviews~\citep{taohuman,yu2024your,rao2025detecting}, where detectors are designed to distinguish machine-written reviews from human-authored ones. However, these approaches rely on textual review content, which is often unavailable in broader judgment settings. In this work, we borrow insights from both fields and propose judgment detection to explore the detectability of LLM-produced judgment, using judgment scores without accessing textual feedback.

\section{Task Statement}

A \emph{judgment} refers to an assessment made over one or more candidates \(c \in \mathcal{C}\), where \(|\mathcal{C}|\) denotes the size of the candidate set. A judgment score is denoted by \(j = (j_{1}, \dots, j_{d}) \in \mathcal{Y}^{d}\). It can be either \emph{single-dimensional} (\(d = 1\)), reflecting an assessment toward a single aspect, or \emph{multi-dimensional} (\(d > 1\)), where each component \(J_{i}\) corresponds to a distinct evaluation aspect (e.g., relevance, fluency, coherence). With these definitions, we formulate the task as follows:

\begin{judgedef}[Judgment Detection]
LLM-generated judgment detection is defined over \emph{judgment groups}. A judgment group is given by $G = \{(c^i, j^i)\}_{i=1}^k$, where each candidate $c^i \in \mathcal{C}$ is paired with a judgment score $j^i \in \mathcal{J}$. The task is to classify whether a group $G$ originates from a human judge or from an LLM. Formally, the label space is $L = \{0,1\}$, where $\ell=0$ denotes human-produced judgments and $\ell=1$ denotes LLM-generated judgments. The goal is to learn a function $f_\theta: G \to [0,1]$, where $f_\theta(G)$ outputs the probability that $G$ was generated by an LLM. The final prediction is obtained as $\hat{y} = \mathbb{I}[f_\theta(G) \geq \tau]$, with threshold $\tau \in [0,1]$ and indicator function $\mathbb{I}[\cdot]$.
\end{judgedef}


When the group size is 1, \textit{i.e.}, $|G| = 1$, the task is degraded to an i.i.d.\ (instance-level) detection setting, where each judgment is treated independently. When $|G| > 1$, the group setting better reflects real practice, since judgments are usually produced in batches (e.g., a reviewer scores multiple papers or an annotator evaluates a set of model outputs), and collective patterns across the group can reveal whether the judgments are human-produced or LLM-generated.

\section{Warm-up Analysis: What Matters for LLM-generated Judgment Detection?}
\label{Warm-up Study: What Makes a Good LLM-generated Judgment Detector?}

\setlength\intextsep{0pt}
\setlength\columnsep{10pt}
\begin{wrapfigure}[11]{r}{3.4in}
    \vspace{-4pt}
    \centering
    \includegraphics[width=3.4in]{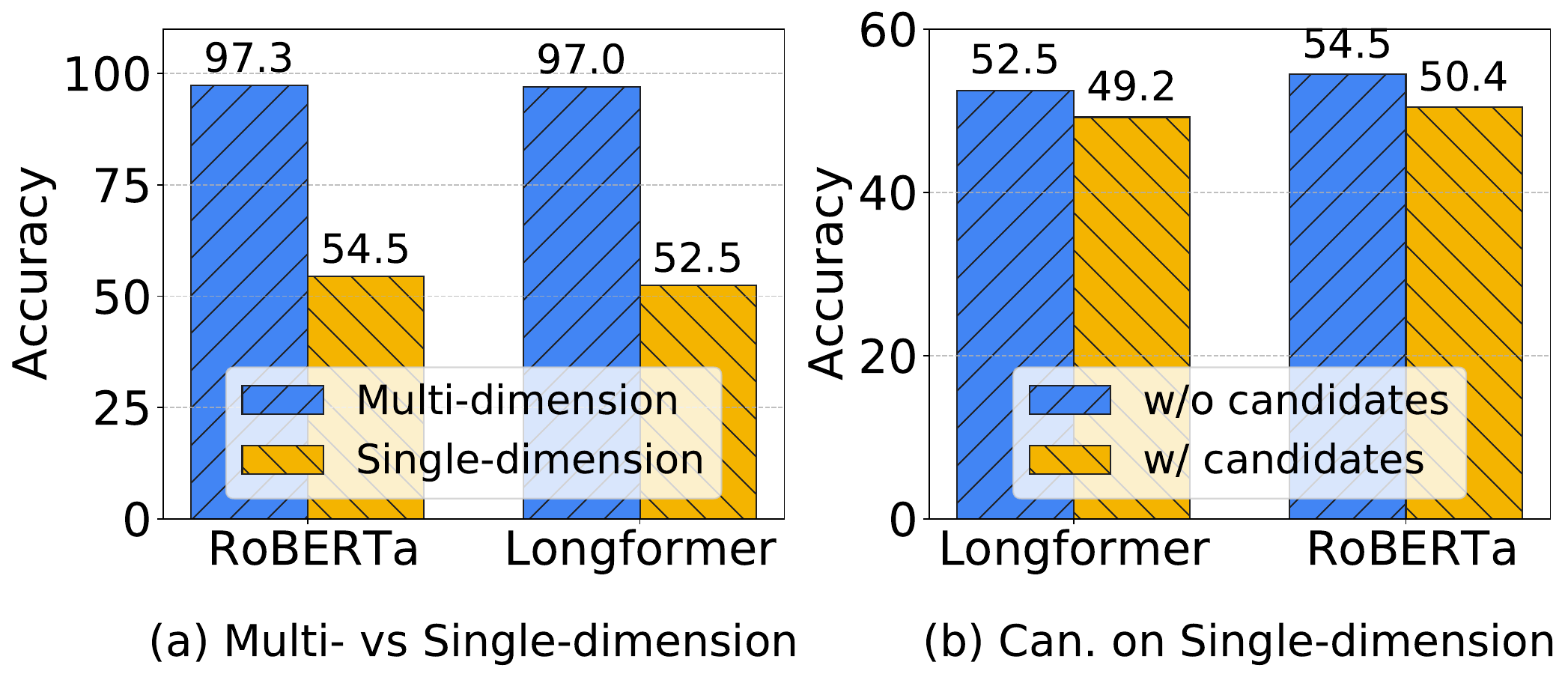}
    \captionsetup{aboveskip=0pt,belowskip=0pt}
    \caption{Multi- vs Single-dimension and Candidate Effect.}
    \label{fig:warm_up}
\end{wrapfigure}

To understand the key ingredients of a reliable judgment detector, we first conduct a warm-up study by adapting LLM-generated text detection methods to the judgment detection setting. Specifically, we employ small language models (SLM)-based detectors~\citep{wu2024detectrl}, \emph{RoBERTa} and \emph{Longformer}, as $f_\theta$ and evaluate them on four datasets: \emph{Helpsteer2}, \emph{Helpsteer3}, \emph{NeurIPS}, and \emph{ANTIQUE}. More information about implementation and dataset can be found in \cref{Experiment Settings}.

\noindent\textbf{Multi-dimension vs Single-dimension performance.}  
As shown in \cref{fig:warm_up} (a), both RoBERTa and Longformer achieve high accuracy in the \emph{multi-dimension} scenarios (Helpsteer2 and NeurIPS) but perform poorly in the \emph{single-dimension} scenarios (Helpsteer3 and ANTIQUE). We assume that this discrepancy arises because, in multi-dimension settings, the detectors can exploit distributional differences in how humans and LLMs assign scores across multiple judgment dimensions, whereas in single-dimensional settings, such distributional cues are almost absent.

\noindent\textbf{Adding candidate information.}  
We further extend the single-dimension setting by providing candidate texts alongside their judgments, exploring whether the detectors can extract and leverage judgment–candidate interaction information. As shown in \cref{fig:warm_up} (b), however, adding candidates does not lead to any performance improvement. This suggests that SLM-based detectors are unable to directly capture and utilize the interaction between judgments and candidate content from raw input.


\noindent\textbf{Takeaway.}  
From this warm-up study, we identify two complementary types of information that a reliable judgment detector should exploit:  
\ding{182} \textbf{Judgment-Intrinsic Features}, revealed by the large performance gap between multi-dimension and single-dimension settings, indicating that distributional patterns within judgment scores themselves can be highly informative; and  
\ding{183} \textbf{Judgment-Candidate Interaction Features}, which capture how judgment scores relate to the underlying candidate content but remain largely unexplored by existing methods. 
These findings highlight that existing SLM-based text detection methods mainly leverage judgment-intrinsic patterns but fail to capture judgment–candidate interactions, which are especially critical in single-dimension scenarios.  

\section{\textit{J-Detector}: A Lightweight and interpretable Detector}

\begin{figure}
    \centering
    \includegraphics[width=1.0\linewidth]{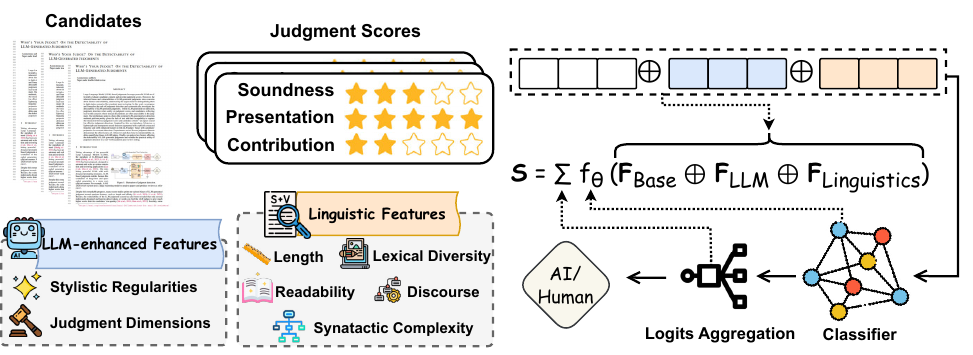}
    \caption{The overview pipeline of our \textit{J-Detector} for LLM-generated judgment detection.}
    \label{fig:pipeline}
\end{figure}

To address the limitation of existing text detectors and design an effective and robust approach for LLM-generated judgment detection, we first identify three criteria that a good LLM-generated judgment detector should embody:
\begin{itemize}[leftmargin=0pt, itemsep=0pt, topsep=0pt, parsep=0pt, partopsep=0pt]
    \item (\textbf{Accurate}) The detection method should be able to leverage both Judgment-Intrinsic Features and Judgment-Candidate Interaction Features to deliver reliable detection results in various scenarios.
    \item (\textbf{Efficient}) Both the training and inference of the detector should incur minimal computational overhead, enabling the method to be deployed in large-scale judgment detection scenarios.
    \item (\textbf{Interpretable}) The detection method should be interpretable to support bias analysis in LLM judges.
\end{itemize}

Following these principles, we design \textit{J-Detector}, an accurate, lightweight and interpretable detector involving the following components. The overview pipeline is presented in \cref{fig:pipeline}.

\noindent\textbf{Feature Augmentation.}
Let \(\mathbf{F}\) denote the instance-level feature vector used by \textit{J-Detector}. We construct it by concatenating three types of features together:
\begin{equation}
\mathbf{F} \;=\; \mathbf{F}_{\text{base}} \;\oplus\; \mathbf{F}_{\text{LLM}} \;\oplus\; \mathbf{F}_{\text{linguistic}},
\label{eq:feature_concat}
\end{equation}
where \(\mathbf{F}_{\text{base}}\) contains the \emph{given judgment scores}. \(\mathbf{F}_{\text{LLM}}\) and \(\mathbf{F}_{\text{linguistic}}\) are \emph{LLM-enhanced features} and \emph{linguistic features} we extract from candidates content, which act as distilled information of candidates and are leveraged to link judgment scores with candidates' content.

\underline{LLM-enhanced Features.}
Borrowing insights from LLM-based text detection methods~\citep{bao2024fast}, we propose LLM-enhanced features to produce the following types of features:
\begin{itemize}[leftmargin=0pt, itemsep=0pt, topsep=0pt, parsep=0pt, partopsep=0pt]
    \item \textit{Stylistic regularities:} scores reflecting surface polish and presentation patterns of the candidates, including \emph{style}, \emph{wording}, and \emph{format}. These aim to capture the spurious preference LLM judges tend to have over superficial attributes~\citep{li2025preference}.
    \item \textit{Judgment-aligned dimensions:} scores aligned to the same dimensions used in the given judgment scores. These aim to enhance features by leveraging the similarity of biases across LLM judges.
\end{itemize}

By injecting these high-level, bias-informed signals, LLM-enhanced Features enable the detector to better capture subtle judgment patterns that are difficult to learn from raw candidate content alone. 

\underline{Linguistic Features.}  
We further introduce linguistic features \(\mathbf{F}_{\text{linguistic}}\) to capture low-level linguistic regularities that often correlate with systematic biases of LLM judges. Specifically, we extract the following aggregated features from the candidate content:

\begin{itemize}[leftmargin=0pt, itemsep=0pt, topsep=0pt, parsep=0pt, partopsep=0pt]
    \item \textit{Length:} total token and character counts, as well as average sentence length, to capture the \emph{length bias} where LLM judges favor lengthy content and responses~\citep{wei2025systematic}.
    \item \textit{Lexical diversity:} unique-token ratio and average word length, which reflect the \emph{surface beauty bias} of LLM-generated judgments compared to human-produced ones~\citep{chen2024humans}.
    \item \textit{Readability:} a composite readability index (e.g., Coleman--Liau), measuring the \emph{fluency bias} where LLMs tend to favor superficially fluent texts, disregarding their true quality~\citep{wu2025style}.
    \item \textit{Syntactic complexity:} dependency tree depth and average dependency distance, used to identify the \emph{complexity bias} often observed in LLM judges~\citep{ye2024justice}.
    \item \textit{Discourse/hedging:} the frequency of discourse markers and hedging expressions, capturing the \emph{presentation bias} of LLM, which prefer content with confident tones~\citep{kharchenko2025think}.
\end{itemize}

These features provide a compact yet informative summary of linguistic cues, enabling the detector to exploit stable and interpretable signals that are complementary to LLM-enhanced features.

\noindent\textbf{Model Training.} Given labeled instances \((\mathbf{F}, y)\), we train a lightweight binary classifier $f_\theta$ (e.g., RandomForest~\citep{breiman2001randomforest}) to output a \emph{logit} \(z \in \mathbb{R}\) indicating the likelihood that the judgment was generated by an LLM (\(y = 1\)) or by a human (\(y = 0\)). The classifier is trained using the augmented feature \(\mathbf{F}\) and serves as the instance-level building block for group-level decisions.

\noindent\textbf{Group-level Aggregation.} To enable the group-level detection setting, we propose a simple aggregation method to produce the group-level label give each single prediction. Given a group \(G\) consisting of \(k\) judgments with instance-level logits \(\{\hat{z}_1,\dots,\hat{z}_k\}\), we aggregate the evidence using sum aggregation: $\mathrm{score}(G) \;=\; \sum_{i=1}^{k} \hat{z}_i$.


In summary, \textit{J-Detector} is designed to satisfy the three criteria identified at the beginning of this section. First, by incorporating both LLM-enhanced and linguistic features, it is able to capture not only Judgment-Intrinsic Features but also critical Judgment--Candidate Interaction Features, enabling accurate detection across single-dimensional and multi-dimensional scenarios. Second, it builds on a lightweight binary classifier, making both training and inference highly efficient and thus suitable for large-scale deployment. Third, since the features are semantically clear and the classifier itself is simple, the framework offers strong interpretability, which can be leveraged to systematically quantify and analyze the biases of LLM judges.

\section{Main Experiment}
\subsection{Experiment Settings}
\label{Experiment Settings}

\noindent\textbf{Datasets.} We build a comprehensive LLM-generated judgment detection dataset, \textit{JD-Bench}, which integrates four representative datasets covering three judgment types: pointwise, pairwise and listwise~\citep{li2024generation}. Among them, \emph{HelpSteer2} provides large-scale pointwise human ratings of LLM responses for helpfulness evaluation, while \emph{HelpSteer3} extends this with pairwise human preference comparisons. The \emph{NeurIPS Review dataset} offers expert peer reviews with multi-dimensional scores such as soundness and novelty, representing high-stakes evaluation. Finally, \emph{ANTIQUE} supplies listwise human judgments for ranking documents in non-factoid question answering. All four datasets contain human-labeled judgments as reliable references, and we further collect LLM-generated judgments from a diverse pool of models. In total, \textit{JD-Bench} covers a wide spectrum of model families, including \emph{OpenAI}, \emph{Anthropic}, and \emph{Google} for closed-source models, and \emph{LLaMA}, \emph{Qwen}, \emph{Mistral}, and \emph{DeepSeek} for open-source models, ensuring diversity in judgment patterns.

\noindent\textbf{Compared Methods.} In our main experiment, we compare our proposed \textit{J-Detector} against a series of baseline methods, all of which are listed as follows:
\begin{itemize}[leftmargin=0pt, itemsep=0pt, topsep=0pt, parsep=0pt, partopsep=0pt]
    \item \textbf{SLM-based Detector.} In line with SLM-based text detectors~\citep{yu2025your}, this approach feeds either the judgment scores alone or the judgment scores together with the candidate content (w/ candidates) to train a small language model-based classifier to predict whether the judgment was produced by a human or from an LLM.

    \item\textbf{LLM-as-a-judge-detector.} Inspired by logits-based detection in AI-generated text detection~\citep{mitchell2023detectgpt}, where a surrogate LLM is used to compute likelihoods, we adopt a single LLM that first generates judgment scores and then compares them with the judgment scores to be detected, making the detection decision based on their similarity.

    \item\textbf{Sample-level LLM-based Analysis.} Inspired by recent agent-based frameworks that maintain guideline banks for distinguishing human and AI text~\citep{li2025agent}, we let the LLM analyze Human–LLM judgment-candidate pairs to extract concise instance-level features (e.g., length bias in LLM judgments), which are stored in a feature bank to capture regularities useful for detection.

    \item\textbf{Distribution-level LLM-based Analysis.} Drawing inspiration from recent work that guides LLMs in structured extraction and analysis of visual summaries~\citep{liu2025can}, we provide the model with dataset-level summaries (e.g., per-label histograms and correlations), enabling it to incorporate global and distributional cues into the detection decision.
\end{itemize}

\noindent\textbf{Implementation Details.}
We implement our \textit{J-Detector} using three models from the Scikit-learn library~\citep{scikit-learn}: LGBM~\citep{ke2017lightgbm}, RandomForest~\citep{breiman2001randomforest}, and XGB~\citep{chen2016xgboost}. We employ \emph{Qwen-3-8B} for both feature augmentation and as the backbone for LLM-based baselines. For SLM-based methods, we use \emph{RoBERTa-base} and \emph{Longformer-4096}. For SLM training, we use a batch size of 8 and fine-tune the SLM for 3 epochs on each dataset. In the main experiments, the group size is fixed to $k$=4. More details, including the \textit{JD-Bench} construction, design of baseline methods, and implementation specifics are provided in \cref{Experiment Implementation Details}.

\subsection{Main Result}
\label{Main Result}

\vspace{0.8em}
\begin{table}[h!]
\scriptsize
\setlength{\tabcolsep}{4pt}
\renewcommand{\arraystretch}{0.95}
\setlength{\abovetopsep}{0pt}
\setlength{\belowrulesep}{0pt}
\setlength{\aboverulesep}{0pt}
\centering
\caption{Main experimental results on \textit{JD-Bench}. We report F1 and AUROC scores, with the best results highlighted in bold. Each experiment is repeated five times, and average scores are reported.}
\label{tab:main_result}
\begin{tabular}{lcccccccccc}
\toprule
\rowcolor{headercolor}
\textbf{Method} & \multicolumn{2}{c}{\textbf{Helpsteer2}} & \multicolumn{2}{c}{\textbf{Helpsteer3}} & \multicolumn{2}{c}{\textbf{NeurIPS}} & \multicolumn{2}{c}{\textbf{ANTIQUE}} & \multicolumn{2}{c}{\textbf{AVG}} \\
\cmidrule(lr){2-11}\addlinespace[2pt]
& F1 & AUROC & F1 & AUROC & F1 & AUROC & F1 & AUROC & F1 & AUROC \\
\midrule

\rowcolor{white}\multicolumn{11}{c}{\textit{\textbf{SLM-based methods}}} \\
\rowcolor{gray!8}RoBERTa & 98.1 & 99.6 & 50.9 & 64.5 & 96.2 & 99.4 & 30.0 & 56.8 & 68.8 & 80.1 \\
\rowcolor{white}RoBERTa w/ candidates& 98.1 & 99.6 & 50.0 & 63.4 & 96.3 & 99.3 & 27.6 & 56.6 & 68.0 & 79.7 \\
\rowcolor{gray!8}Longformer& 98.1 & 99.7 & 54.5 & 65.7 & 96.2 & 99.5 & 30.6 & 56.6 & 69.9 & 80.4 \\
\rowcolor{white}Longformer w/ candidates & 98.1 & 99.7 & 51.4 & 64.3 & 96.2 & 99.4 & 21.8 & 48.8 & 66.9 & 78.0 \\
\midrule

\rowcolor{white}\multicolumn{11}{c}{\textit{\textbf{LLM-based methods}}} \\
\rowcolor{gray!8}LLM& 51.5 & 50.3 & 50.3 & 50.1 & 43.9 & 50.2 & 49.6 & 49.9 & 48.8 & 50.1 \\
\rowcolor{white}LLM w/ Sample-level& 49.8 & 49.7 & 49.6 & 50.2 & 50.5 & 50.4 & 50.9 & 50.3 & 50.2 & 50.2 \\
\rowcolor{gray!8}LLM w/ Distribution-level& 52.1 & 50.0 & 48.8 & 50.3 & 49.6 & 49.8 & 50.7 & 50.1 & 50.3 & 50.1 \\
\rowcolor{white}LLM w/ Sample-level + Distribution-level & 58.7 & 50.4 & 49.4 & 49.6 & 51.2 & 50.2 & 50.2 & 49.9 & 52.4 & 50.0 \\
\midrule

\rowcolor{white}\multicolumn{11}{c}{\textit{J-Detector (ours)}} \\
\rowcolor{gray!8}LGBM & 99.6 & \bf{100.0} & 68.1 & 73.3 & \bf{98.7} & \bf{99.9} & \bf{85.4} & \bf{93.3} & 88.0 & 91.6 \\
\rowcolor{white}RandomForest& 99.5 & \bf{100.0} & \bf{74.0} & \bf{77.0} & 97.0 & 99.7 & 82.6 & 90.6 & \bf{88.3} & \bf{91.8} \\
\rowcolor{gray!8}XGB& \bf{99.8} & \bf{100.0} & 68.5 & 73.6 & 98.4 & 99.8 & 84.2 & 92.3 & 87.7 & 91.4 \\
\bottomrule
\end{tabular}

\vspace{0.6em}
\end{table}

\noindent\textbf{SLM-based Methods Analysis.} As we discussed in \cref{Warm-up Study: What Makes a Good LLM-generated Judgment Detector?}, SLM-based methods perform strongly on multi-dimensional datasets like Helpsteer2 (98.1\% F1 on RoBERTa) and NeurIPS (96.2\% on RoBERTa), but drop to around 50–55\% F1 on single-dimensional datasets like Helpsteer3 and Antique. Even adding candidates barely helps. This shows SLMs rely on inter-dimension patterns and fail to link judgments with candidates when such distributional cues are absent.

\noindent\textbf{LLM-based Methods Analysis.} Furthermore, all LLM-based methods hover near 50\% F1 score across datasets, indicating almost random guessing. When combining with sample-level comparative analysis and distribution-level chart reasoning, LLM-based detection methods yield some gains in multi-dimensional datasets (e.g., from 51.5\% to 58.7\% F1 score). While this improvement doesn't appear in Helpsteer3 and ANTIQUE, we conclude that LLM-based detectors also suffer from leveraging judgments-candidates interaction, with either sample- or distribution-level methods.

\noindent\textbf{J-Detector Analysis.} Compared with them, \textit{J-Detector} achieves the best detection performance across all 4 datasets and 2 metrics, far surpassing all baselines. Noted that in the single-dimensional judgment scenarios, \textit{J-Detector} yields much better detection performance compared with other baselines. This demonstrates that explicitly modeling the distributional patterns and biases of LLM judgments is crucial for accurate detection, enabling robust performance in both single-dimensional and multi-dimensional judgment detection scenarios.

\begin{wrapfigure}{!th}{0.7\textwidth}
    \centering
    \includegraphics[width=\linewidth]{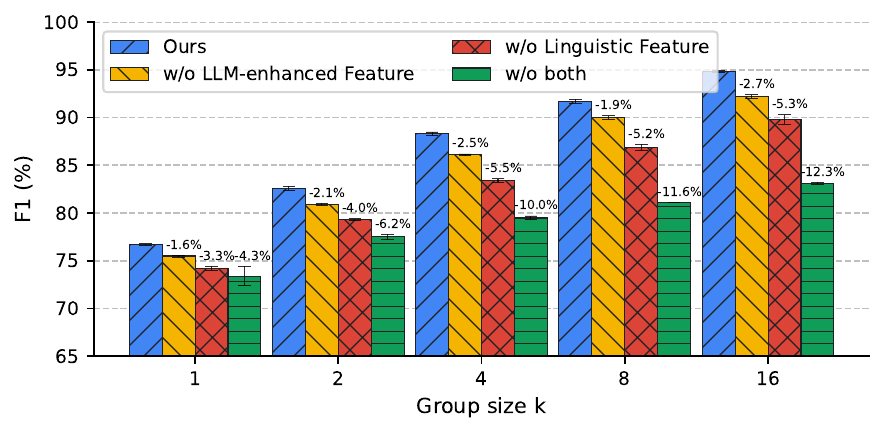}
    \caption{Ablation study on LLM-enhanced and linguistic features.}
    \label{fig:ablation_study}
\end{wrapfigure}

\noindent\textbf{Ablation Study.} \cref{fig:ablation_study} shows that both LLM-enhanced and linguistic features consistently improve performance across all group sizes. Removing either feature causes the F1 score to drop at every group size---for example, at k = 16, removing linguistic features lowers F1 by 5.3\%, and removing both leads to a 12.3\% drop. This demonstrates that the two augmented features are complementary and beneficial across all datasets and group-size settings.

\begin{figure}[h!]
    \centering
    \includegraphics[width=0.49\linewidth]{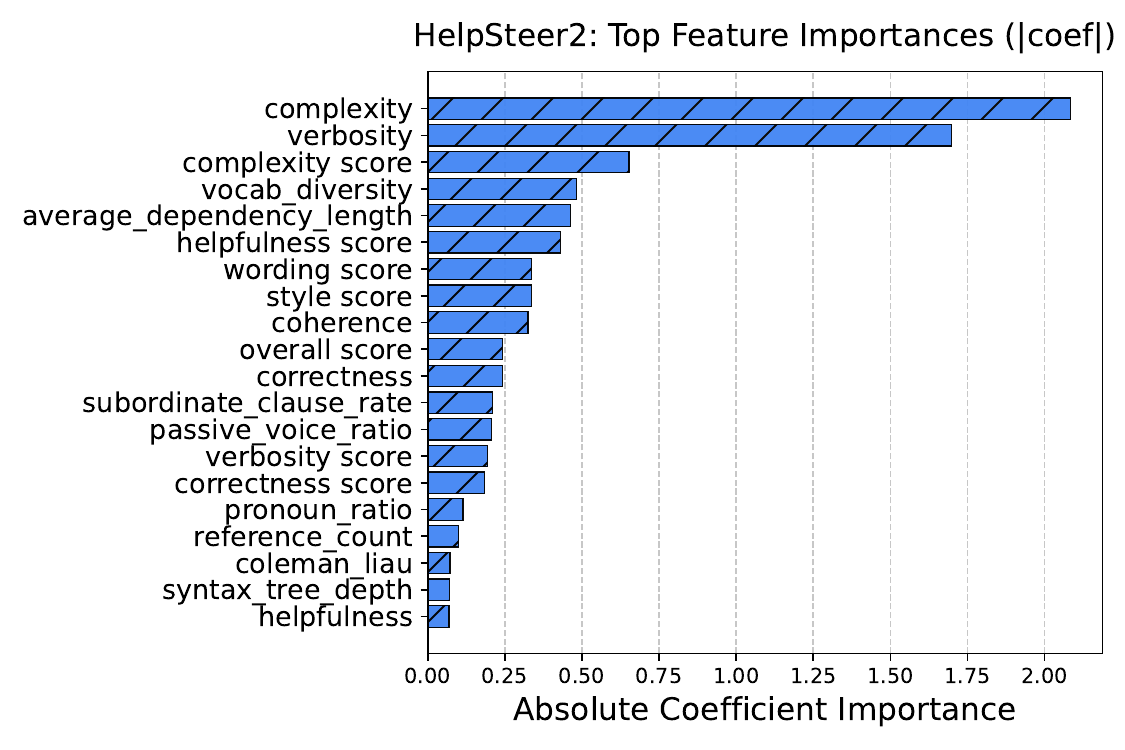}
    \includegraphics[width=0.49\linewidth]{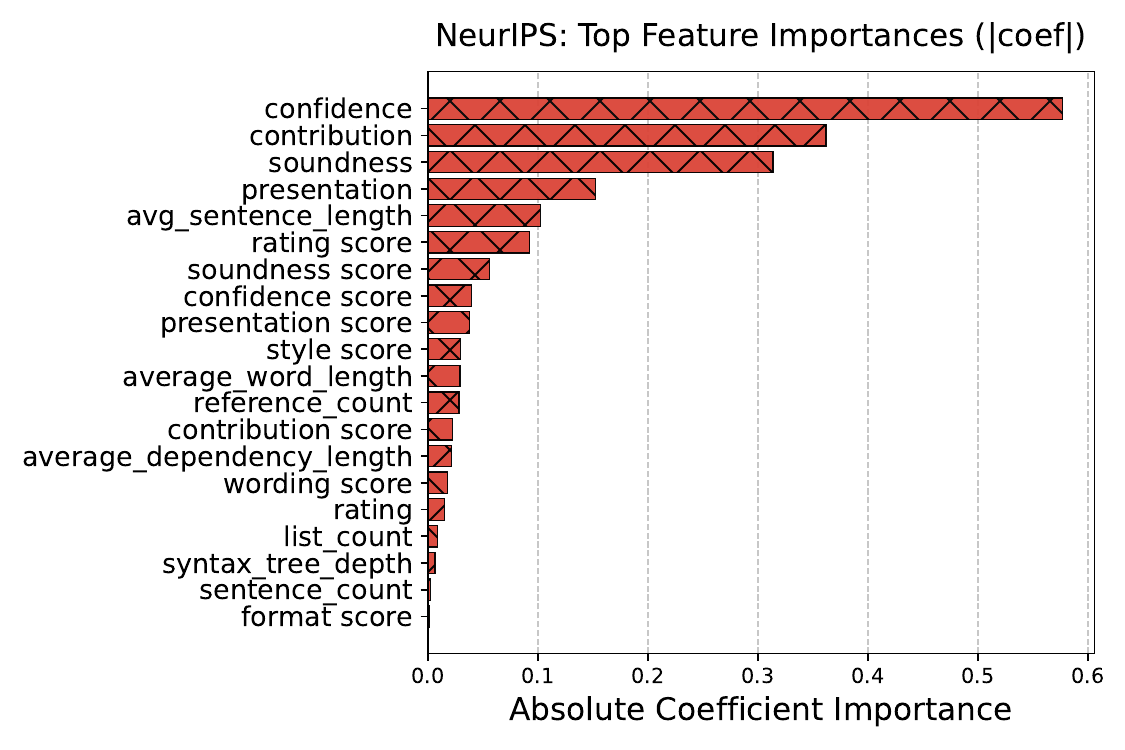}    
    \caption{LLM-as-a-judge bias quantification on Helpsteer2 and NeurIPS.}
    \label{fig:explainability}
\end{figure}

\noindent\textbf{Bias Quantification with J-Detector.}
Additionally, we illustrate how the transparency and interpretability of \textit{J-Detector} can be leveraged to quantify biases in LLM-as-a-judge by analyzing which features most strongly influence the detector’s decisions. Specifically, we select the top 20 most important features ranked by their absolute coefficient values, and report the results on the Helpsteer2 and NeurIPS datasets in \cref{fig:explainability}. The analysis reveals that base judgment score features provide strong signals for distinguishing LLM-generated judgments from human-produced ones, highlighting the critical role of \textit{Judgment-Intrinsic Features}. As shown in the figure, LLM judges exhibit the strongest bias in the \textit{complexity} and \textit{confidence} dimensions for the two datasets, respectively, consistent with prior findings that LLMs tend to favor more complex responses~\citep{ye2024justice,yang2025quantifying} and often display overconfidence~\citep{kadavath2022language}. In addition, we observe common cross-dataset biases such as \textit{length bias} (captured by \texttt{average\_dependency\_length}) and \textit{beauty bias} (reflected in style-related scores), which echo broader concerns about spurious preference and correlations in LLM-based judgments~\citep{wang2023adversarial,shi2024optimization}.



\section{Further Analysis}

In this section, we empirically analyze the key factors that influence the detectability of the LLM-generated judgment, as well as present a real-world application to combine LLM-based judgment detection with text detection in real-world academic peer reviewing scenarios.



\subsection{Detectability Analysis}

\begin{figure}[h!]
    \centering
    \includegraphics[width=1.0\linewidth]{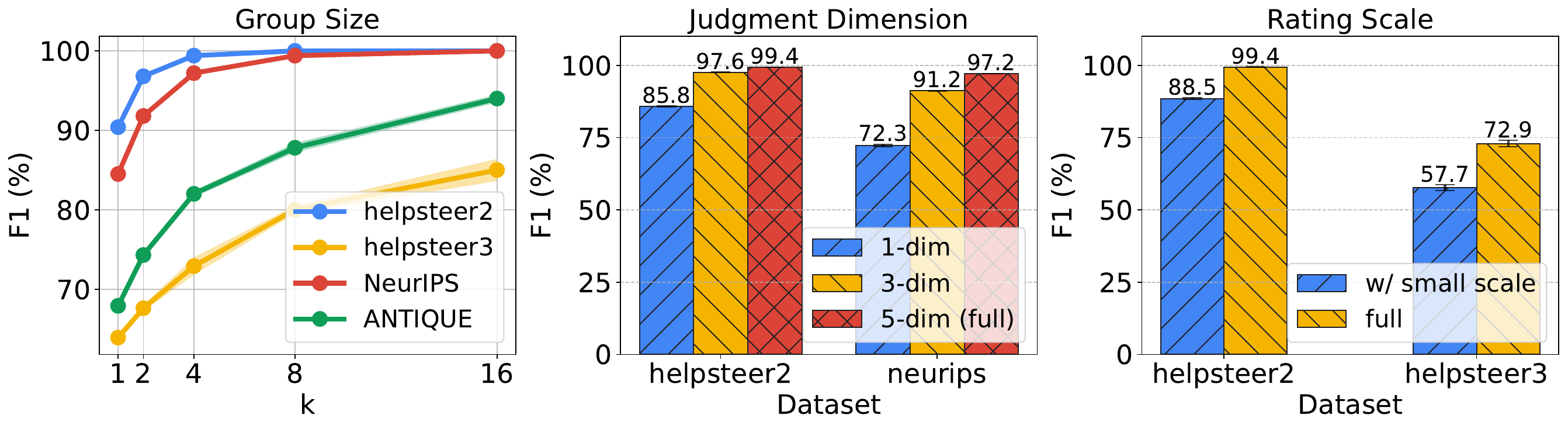}
    \caption{Detectability analysis on group size, judgment dimensions and rating scale.}
    \label{fig:detectability}
\end{figure}

\noindent\textbf{Detectability analysis across group size, judgment dimensions, and rating scale.} 
\cref{fig:detectability} shows that group size is a key factor in the detectability of LLM-generated judgments: the F1 score consistently improves as the group size increases across all four datasets (e.g., F1 score in Helpsteer3 rises from 63.9\% at $k=1$ to 85.0\% at $k=16$). 
The number of judgment dimensions also plays an important role; for instance, when only a single dimension out of the five is used in the NeurIPS dataset, the F1 score drops substantially (from 97.2\% to 72.3\%). This confirms that multi-dimensional judgments provide richer distributional signals as Judgment-Intrinsic Features for detection. 
\begin{figure}[h!]
    \centering
    \includegraphics[width=1.0\linewidth]{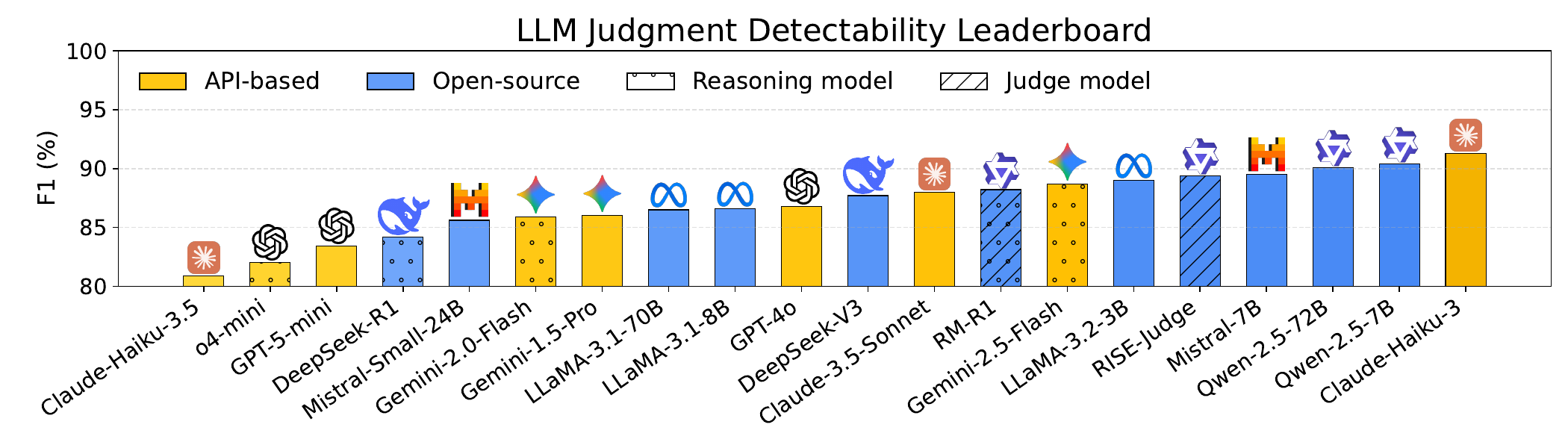}
    \caption{Detectability leaderboard on 20 LLMs. RM-R1 and RISE-Judge are based on Qwen-2.5-7B.}
    \label{fig:leaderboard}
\end{figure}

\begin{wrapfigure}{!th}{0.55\textwidth}
    \centering
    \includegraphics[width=\linewidth]{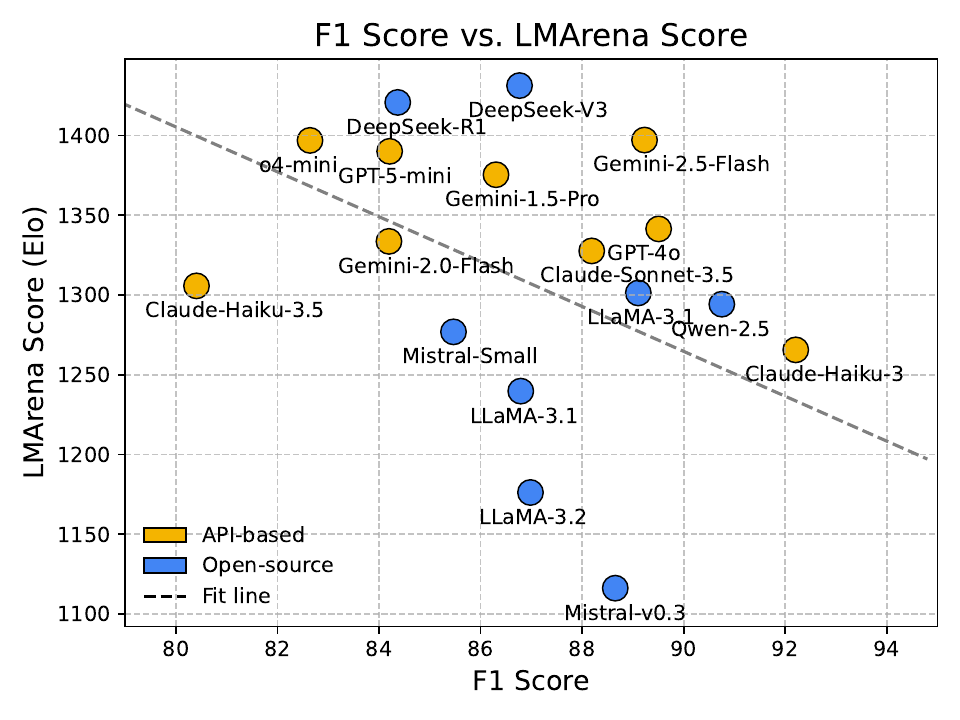}
    \caption{Correlation between judge LLMs' detectability and LMArena score.}
    \label{fig:arena}
\end{wrapfigure}
In addition, the granularity of the rating scale further impacts detectability: collapsing to a coarse scale (e.g., merging $-3/-2/-1$ into $-1$ and $1/2/3$ into $1$ in Helpsteer3) leads to degraded performance (e.g., F1 drops from 72.9\% to 57.7\%). 
Overall, these results underscore that group size, the number of dimensions, and the rating scale collectively shape how detectable LLM-generated judgments are.

\noindent\textbf{Detectability of Various LLM Judges.} 
Additionally, \cref{fig:leaderboard} summarizes the detectability leaderboard across 20 LLMs, averaged over different group sizes. 
We observe that API-based models (yellow bars) are generally more difficult to detect than open-source models (blue bars), indicating that closed commercial systems such as GPT-5-mini and Claude-Haiku-3 produce judgments that more closely resemble human annotations.

Within the same model families, larger models tend to be less detectable than smaller ones: for instance, among LLaMA-3 and Qwen-2.5 families, larger models consistently achieve lower detectability. 
Moreover, reasoning models (dotted bars) and specialized judge models (striped bars) consistently achieve higher robustness than standard LLMs, suggesting that models explicitly optimized for reasoning or evaluation align more closely with human judgment distributions and are therefore harder to distinguish from human judges.

As presented in \cref{fig:arena}, we also study the correlation between the detectability of different LLM judges and their LMArena score~\citep{chiang2024chatbot}, which is a proxy of LLMs' alignment degree with human preference and value. We find a clear negative correlation: models with higher alignment scores are systematically less detectable. This observation reinforces our previous findings, supporting the hypothesis that as models become better aligned with human values, the gap between their judgments and human annotations narrows, making their outputs increasingly difficult to distinguish from those of human judges.

For LLM-generated judgment detectability, we also theoretically prove and demonstrate each influence factor's effect and put it in Appendix~\ref{Theoretically Analysis on LLM-generated Judgment Detectability}.

\subsection{Judgment Detection with Multiple LLM Judges}
\begin{wrapfigure}{!th}{0.4\textwidth}
    \centering
    \includegraphics[width=\linewidth]{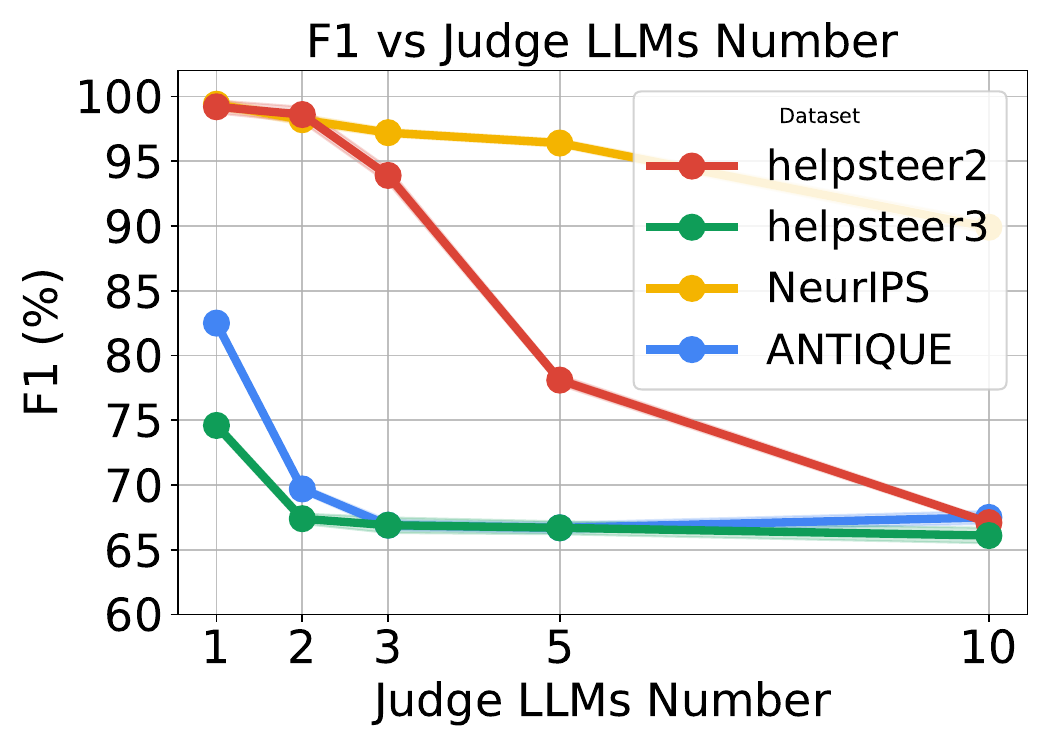}
    \caption{Detectability of LLM-generated judgment in multiple LLM judges setting.}
    \label{fig:model_num}
\end{wrapfigure}
In this section, we examine how the detectability of LLM-generated judgments changes when multiple LLM judges are involved. This setting reflects real-world scenarios where judgments may come from a diverse pool of LLMs. As shown in \cref{fig:model_num}, we randomly sample 2, 3, 5, or 10 LLMs from our \textit{JD-Bench} and mix their judgments in both the training and testing sets. We observe a substantial drop in detection performance across all four datasets (e.g., the F1 score decreases from 99.8\% to 66.9\% on Helpsteer2). This suggests that detecting LLM-generated judgments becomes significantly more challenging when multiple LLM judges are present, as detectors must learn to recognize distinct patterns from different models. Notably, the performance drop is relatively small on the NeurIPS dataset, indicating stronger shared biases among LLM judges in that domain. One promising direction for future work is to explore effective LLM-generated judgment detection methods under multiple judges' settings. 

\subsection{Judgment-Text Co-Detection: An Application}


In this section, we explore two real-world scenarios where LLM-generated judgment detection can support peer review authenticity checking. First, the few-shot detection setting simulates cases where a new conference is launched or the review form has changed. Here, we set the number of training samples to be 60. Second, the missing-text detection setting addresses the common case where reviews lack enough textual feedback. We simulate this setting by masking 15\% of the text reviews.

\setlength\intextsep{0pt}
\setlength{\aboverulesep}{0pt}
\setlength{\belowrulesep}{0pt}
\begin{wraptable}[9]{r}{3.9in}
    \centering
    \caption{An application to leverage judgment and text feedback for AI-generated review detection in few-shot and missing review scenarios.}
    \label{fig:application}
    \small
    \begin{tabular}{l c c}
        \toprule 
        \rowcolor{headercolor} \textbf{Method} & \textbf{Few-shot} & \textbf{Missing review} \\
        \midrule 
        w/ RoBERTa-text & 67.2 & 90.5 \\
        \rowcolor{gray!8} w/ \textit{J-Detector} & 64.4 & 86.2 \\
        w/ RoBERTa-text \& \textit{J-Detector} & \bf{74.6} & \bf{99.3} \\
        \bottomrule
    \end{tabular}
\end{wraptable}
The results in \cref{fig:application} show that combining the \textit{J-Detector} with a text-based detector (RoBERTa-text) achieves the best performance in both settings (74.6\% vs. 67.2\% in few-shot, and 99.3\% vs. 90.5\% in missing-text), outperforming either method alone. This demonstrates that LLM-generated judgment detection provides complementary signals to text-based detectors and is highly valuable in real-world low-resource or judgment score-only scenarios for robust and reliable detection.

\section{Conclusion}

In this work we introduced judgment detection as the task of distinguishing human from LLM-generated judgments and proposed \textit{J-Detector}, a lightweight, interpretable detector enhanced with linguistic and LLM-based features. Experiments on \textit{JD-Bench} show that \textit{J-Detector} consistently outperforms baselines, while our theoretical and empirical analyses reveal that detectability improves with larger group size, richer dimensions, finer rating scales, and greater human–LLM divergence. Using \textit{J-Detector}’s transparency, we further quantified systematic biases in LLM judges, such as complexity, confidence, and length biases, and demonstrated practical value in peer-review authenticity checking. These findings establish LLM-generated judgment detection as a key safeguard for ensuring fairness and accountability in LLM-as-a-judge systems.

\section*{Ethics Statement}
We adhere to the ICLR Code of Ethics. No private, sensitive, or personally identifiable data are involved. Our work does not raise foreseeable ethical concerns or produce harmful societal outcomes.

\section*{Reproducibility Statement}
Reproducibility is central to our work. All datasets used in our experiments are standard benchmarks that are publicly available. We provide full details of the training setup, model architectures, and evaluation metrics in the main paper and appendix. Upon acceptance, we will release our codebase, including scripts for preprocessing, training, and evaluation, along with configuration files and documentation to facilitate exact reproduction of our results. Random seeds and hyperparameters will also be included to further ensure reproducibility.

\bibliographystyle{abbrvnat}
\nobibliography*
\bibliography{custom}

\begin{thebibliography}{55}
\providecommand{\natexlab}[1]{#1}
\providecommand{\url}[1]{\texttt{#1}}
\expandafter\ifx\csname urlstyle\endcsname\relax
  \providecommand{\doi}[1]{doi: #1}\else
  \providecommand{\doi}{doi: \begingroup \urlstyle{rm}\Url}\fi

\bibitem[Bao et~al.(2024)Bao, Zhao, Teng, Yang, and Zhang]{bao2024fast}
G.~Bao, Y.~Zhao, Z.~Teng, L.~Yang, and Y.~Zhang.
\newblock Fast-detectgpt: Efficient zero-shot detection of machine-generated text via conditional probability curvature.
\newblock In \emph{ICLR}, 2024.

\bibitem[Beigi et~al.(2024)Beigi, Tan, Mudiam, Chen, Shu, and Liu]{beigi2024model}
A.~Beigi, Z.~Tan, N.~Mudiam, C.~Chen, K.~Shu, and H.~Liu.
\newblock Model attribution in llm-generated disinformation: A domain generalization approach with supervised contrastive learning.
\newblock In \emph{2024 IEEE 11th International Conference on Data Science and Advanced Analytics (DSAA)}, pages 1--10. IEEE, 2024.

\bibitem[Breiman(2001)]{breiman2001randomforest}
L.~Breiman.
\newblock Random forests.
\newblock \emph{Machine learning}, 45\penalty0 (1):\penalty0 5--32, 2001.

\bibitem[Chang et~al.(2025)Chang, Li, Zhang, Kong, Wu, Guo, and Wong]{chang2025treereview}
Y.~Chang, Z.~Li, H.~Zhang, Y.~Kong, Y.~Wu, Z.~Guo, and N.~Wong.
\newblock Treereview: A dynamic tree of questions framework for deep and efficient llm-based scientific peer review.
\newblock \emph{arXiv preprint arXiv:2506.07642}, 2025.

\bibitem[Chen et~al.(2024)Chen, Chen, Liu, Jiang, and Wang]{chen2024humans}
G.~Chen, S.~Chen, Z.~Liu, F.~Jiang, and B.~Wang.
\newblock Humans or llms as the judge? a study on judgement bias.
\newblock In \emph{Proceedings of the 2024 Conference on Empirical Methods in Natural Language Processing}, pages 8301--8327, 2024.

\bibitem[Chen and Guestrin(2016)]{chen2016xgboost}
T.~Chen and C.~Guestrin.
\newblock Xgboost: A scalable tree boosting system.
\newblock In \emph{Proceedings of the 22nd ACM SIGKDD International Conference on Knowledge Discovery and Data Mining}, pages 785--794. ACM, 2016.

\bibitem[Chiang et~al.(2024)Chiang, Zheng, Sheng, Angelopoulos, Li, Li, Zhu, Zhang, Jordan, Gonzalez, et~al.]{chiang2024chatbot}
W.-L. Chiang, L.~Zheng, Y.~Sheng, A.~N. Angelopoulos, T.~Li, D.~Li, B.~Zhu, H.~Zhang, M.~Jordan, J.~E. Gonzalez, et~al.
\newblock Chatbot arena: An open platform for evaluating llms by human preference.
\newblock In \emph{Forty-first International Conference on Machine Learning}, 2024.

\bibitem[Gao et~al.(2023)Gao, Ruan, Sun, Yin, Yang, and Wan]{gao2023human}
M.~Gao, J.~Ruan, R.~Sun, X.~Yin, S.~Yang, and X.~Wan.
\newblock Human-like summarization evaluation with chatgpt.
\newblock \emph{arXiv preprint arXiv:2304.02554}, 2023.

\bibitem[Gehrmann et~al.(2019)Gehrmann, Strobelt, and Rush]{gehrmann2019gltr}
S.~Gehrmann, H.~Strobelt, and A.~M. Rush.
\newblock Gltr: Statistical detection and visualization of generated text.
\newblock In \emph{Proceedings of the 57th Annual Meeting of the Association for Computational Linguistics: System Demonstrations}, pages 111--116. Association for Computational Linguistics, 2019.
\newblock URL \url{https://aclanthology.org/P19-3019}.

\bibitem[Hashemi et~al.(2020)Hashemi, Aliannejadi, Zamani, and Croft]{hashemi2020antique}
H.~Hashemi, M.~Aliannejadi, H.~Zamani, and W.~B. Croft.
\newblock Antique: A non-factoid question answering benchmark.
\newblock In \emph{European Conference on Information Retrieval}, pages 166--173. Springer, 2020.

\bibitem[Hu et~al.()Hu, Ren, Hu, Lin, Wang, Tan, Lyu, Zhang, Xiong, and Wang]{hueditable}
L.~Hu, C.~Ren, Z.~Hu, H.~Lin, C.-L. Wang, Z.~Tan, W.~Lyu, J.~Zhang, H.~Xiong, and D.~Wang.
\newblock Editable concept bottleneck models.
\newblock In \emph{Forty-second International Conference on Machine Learning}.

\bibitem[Hu et~al.(2024)Hu, Liu, Yang, Chen, Tan, Ali, Li, and Wang]{hu2024understanding}
L.~Hu, L.~Liu, S.~Yang, X.~Chen, Z.~Tan, M.~A. Ali, M.~Li, and D.~Wang.
\newblock Understanding reasoning in chain-of-thought from the hopfieldian view.
\newblock \emph{arXiv preprint arXiv:2410.03595}, 2024.

\bibitem[Ippolito et~al.(2020)Ippolito, Duckworth, Callison-Burch, and Eck]{ippolito2020automatic}
D.~Ippolito, D.~Duckworth, C.~Callison-Burch, and D.~Eck.
\newblock Automatic detection of generated text is easiest when humans are fooled.
\newblock In \emph{Proceedings of the 58th Annual Meeting of the Association for Computational Linguistics}, pages 1808--1822. Association for Computational Linguistics, 2020.
\newblock URL \url{https://aclanthology.org/2020.acl-main.164}.

\bibitem[Jeong et~al.(2024)Jeong, Jiang, Tan, Bernard, and Liu]{jeong2024bluetempnet}
U.~Jeong, B.~Jiang, Z.~Tan, R.~Bernard, and H.~Liu.
\newblock Bluetempnet: A temporal multi-network dataset of social interactions in bluesky social.
\newblock \emph{IEEE Data Descriptions}, 2024.

\bibitem[Jiang et~al.(2024)Jiang, Li, Tan, Zhou, Rao, Lerman, Bernard, and Liu]{jiang2024assessing}
B.~Jiang, D.~Li, Z.~Tan, X.~Zhou, A.~Rao, K.~Lerman, H.~R. Bernard, and H.~Liu.
\newblock Assessing the impact of conspiracy theories using large language models.
\newblock \emph{arXiv preprint arXiv:2412.07019}, 2024.

\bibitem[Jin et~al.(2024)Jin, Zhao, Wang, Chen, Zhu, Xiao, and Wang]{jin2024agentreview}
Y.~Jin, Q.~Zhao, Y.~Wang, H.~Chen, K.~Zhu, Y.~Xiao, and J.~Wang.
\newblock Agentreview: Exploring peer review dynamics with llm agents.
\newblock In \emph{EMNLP}, 2024.

\bibitem[Kadavath et~al.(2022)Kadavath, Lin, Ganguli, Askell, Bai, Chen, Goldie, Jones, Joseph, Krueger, Nisan, Amodei, Brown, Olsson, Kaplan, Clark, Christiano, Leike, and Cotra]{kadavath2022language}
S.~Kadavath, A.~Z. Lin, D.~Ganguli, A.~Askell, Y.~Bai, A.~Chen, A.~Goldie, A.~Jones, N.~S. Joseph, D.~Krueger, S.~M. Nisan, D.~Amodei, T.~B. Brown, C.~Olsson, J.~Kaplan, J.~Clark, P.~Christiano, J.~Leike, and A.~Cotra.
\newblock Language models (mostly) know what they know.
\newblock In \emph{Advances in Neural Information Processing Systems (NeurIPS)}, 2022.

\bibitem[Ke et~al.(2017)Ke, Meng, Finley, Wang, Chen, Ma, Ye, and Liu]{ke2017lightgbm}
G.~Ke, Q.~Meng, T.~Finley, T.~Wang, W.~Chen, W.~Ma, Q.~Ye, and T.-Y. Liu.
\newblock Lightgbm: A highly efficient gradient boosting decision tree.
\newblock In \emph{Advances in Neural Information Processing Systems}, 2017.

\bibitem[Kharchenko et~al.(2025)Kharchenko, Roosta, Chadha, and Shah]{kharchenko2025think}
J.~Kharchenko, T.~Roosta, A.~Chadha, and C.~Shah.
\newblock I think, therefore i am under-qualified? a benchmark for evaluating linguistic shibboleth detection in llm hiring evaluations.
\newblock \emph{arXiv preprint arXiv:2508.04939}, 2025.

\bibitem[Lee et~al.()Lee, Phatale, Mansoor, Lu, Mesnard, Ferret, Bishop, Hall, Carbune, and Rastogi]{lee2023rlaif}
H.~Lee, S.~Phatale, H.~Mansoor, K.~R. Lu, T.~Mesnard, J.~Ferret, C.~Bishop, E.~Hall, V.~Carbune, and A.~Rastogi.
\newblock Rlaif: Scaling reinforcement learning from human feedback with ai feedback.

\bibitem[Li et~al.(2024)Li, Jiang, Huang, Beigi, Zhao, Tan, Bhattacharjee, Jiang, Chen, Wu, et~al.]{li2024generation}
D.~Li, B.~Jiang, L.~Huang, A.~Beigi, C.~Zhao, Z.~Tan, A.~Bhattacharjee, Y.~Jiang, C.~Chen, T.~Wu, et~al.
\newblock From generation to judgment: Opportunities and challenges of llm-as-a-judge.
\newblock \emph{arXiv preprint arXiv:2411.16594}, 2024.

\bibitem[Li et~al.(2025{\natexlab{a}})Li, Sun, Huang, Zhong, Jiang, Han, Zhang, Wang, and Liu]{li2025preference}
D.~Li, R.~Sun, Y.~Huang, M.~Zhong, B.~Jiang, J.~Han, X.~Zhang, W.~Wang, and H.~Liu.
\newblock Preference leakage: A contamination problem in llm-as-a-judge.
\newblock \emph{arXiv preprint arXiv:2502.01534}, 2025{\natexlab{a}}.

\bibitem[Li et~al.(2025{\natexlab{b}})Li, Tan, Qian, Li, Chaudhary, Hu, and Shen]{li2025smoa}
D.~Li, Z.~Tan, P.~Qian, Y.~Li, K.~Chaudhary, L.~Hu, and J.~Shen.
\newblock Smoa: Improving multi-agent large language models with sparse mixture-of-agents.
\newblock In \emph{Pacific-Asia Conference on Knowledge Discovery and Data Mining}, pages 54--65. Springer, 2025{\natexlab{b}}.

\bibitem[Li et~al.(2025{\natexlab{c}})Li, Ye, Peng, Yin, and Wan]{li2025agent}
J.~Li, M.~Ye, C.~Peng, X.~Yin, and X.~Wan.
\newblock Agent-x: Adaptive guideline-based expert network for threshold-free ai-generated text detection.
\newblock \emph{arXiv preprint arXiv:2505.15261}, 2025{\natexlab{c}}.

\bibitem[Liang et~al.(2024)Liang, Izzo, Zhang, Lepp, Cao, Zhao, Chen, Ye, Liu, Huang, et~al.]{liang2024monitoring}
W.~Liang, Z.~Izzo, Y.~Zhang, H.~Lepp, H.~Cao, X.~Zhao, L.~Chen, H.~Ye, S.~Liu, Z.~Huang, et~al.
\newblock Monitoring ai-modified content at scale: a case study on the impact of chatgpt on ai conference peer reviews.
\newblock In \emph{Proceedings of the 41st International Conference on Machine Learning}, pages 29575--29620, 2024.

\bibitem[Liu et~al.(2025)Liu, Kamarthi, Zhao, Xu, Wang, Wen, Hartvigsen, Wang, and Prakash]{liu2025can}
H.~Liu, H.~Kamarthi, Z.~Zhao, S.~Xu, S.~Wang, Q.~Wen, T.~Hartvigsen, F.~Wang, and B.~A. Prakash.
\newblock How can time series analysis benefit from multiple modalities? a survey and outlook.
\newblock \emph{arXiv preprint arXiv:2503.11835}, 2025.

\bibitem[Mao et~al.(2025)Mao, Wei, Chen, Fang, and Chau]{mao2025watermarking}
M.~Mao, D.~Wei, Z.~Chen, X.~Fang, and M.~Chau.
\newblock Watermarking large language models: An unbiased and low-risk method.
\newblock In \emph{Proceedings of the 63rd Annual Meeting of the Association for Computational Linguistics (Volume 1: Long Papers)}, pages 7939--7960, 2025.

\bibitem[Mitchell et~al.(2023)Mitchell, Lee, Khazatsky, Manning, and Finn]{mitchell2023detectgpt}
E.~Mitchell, Y.~Lee, A.~Khazatsky, C.~D. Manning, and C.~Finn.
\newblock Detectgpt: Zero-shot machine-generated text detection using probability curvature.
\newblock \emph{arXiv preprint arXiv:2301.11305}, 2023.
\newblock URL \url{https://arxiv.org/abs/2301.11305}.

\bibitem[Pedregosa et~al.(2011)Pedregosa, Varoquaux, Gramfort, Michel, Thirion, Grisel, Blondel, Prettenhofer, Weiss, Dubourg, Vanderplas, Passos, Cournapeau, Brucher, Perrot, and Duchesnay]{scikit-learn}
F.~Pedregosa, G.~Varoquaux, A.~Gramfort, V.~Michel, B.~Thirion, O.~Grisel, M.~Blondel, P.~Prettenhofer, R.~Weiss, V.~Dubourg, J.~Vanderplas, A.~Passos, D.~Cournapeau, M.~Brucher, M.~Perrot, and E.~Duchesnay.
\newblock Scikit-learn: Machine learning in {P}ython.
\newblock \emph{Journal of Machine Learning Research}, 12:\penalty0 2825--2830, 2011.

\bibitem[Rahmani et~al.(2024)Rahmani, Yilmaz, Craswell, Mitra, Thomas, Clarke, Aliannejadi, Siro, and Faggioli]{rahmani2024llmjudge}
H.~A. Rahmani, E.~Yilmaz, N.~Craswell, B.~Mitra, P.~Thomas, C.~L. Clarke, M.~Aliannejadi, C.~Siro, and G.~Faggioli.
\newblock Llmjudge: Llms for relevance judgments.
\newblock In \emph{LLM4Eval@ SIGIR}, 2024.

\bibitem[Rao et~al.(2025)Rao, Kumar, Lakkaraju, and Shah]{rao2025detecting}
V.~Rao, A.~Kumar, H.~Lakkaraju, and N.~B. Shah.
\newblock Detecting llm-generated peer reviews.
\newblock \emph{arXiv preprint arXiv:2503.15772}, 2025.

\bibitem[Shi et~al.(2024)Shi, Yuan, Liu, Huang, Zhou, Sun, and Gong]{shi2024optimization}
J.~Shi, Z.~Yuan, Y.~Liu, Y.~Huang, P.~Zhou, L.~Sun, and N.~Z. Gong.
\newblock Optimization-based prompt injection attack to llm-as-a-judge.
\newblock In \emph{Proceedings of the 2024 on ACM SIGSAC Conference on Computer and Communications Security}, pages 660--674, 2024.

\bibitem[Sun and Lv(2025)]{sun2025zero}
J.~Sun and Z.~Lv.
\newblock Zero-shot detection of llm-generated text via text reorder.
\newblock \emph{Neurocomputing}, 631:\penalty0 129829, 2025.

\bibitem[Tan et~al.(2025)Tan, Yan, Hsu, Han, Wang, Le, Song, Chen, Palangi, Lee, et~al.]{tan2025prospect}
Z.~Tan, J.~Yan, I.~Hsu, R.~Han, Z.~Wang, L.~T. Le, Y.~Song, Y.~Chen, H.~Palangi, G.~Lee, et~al.
\newblock In prospect and retrospect: Reflective memory management for long-term personalized dialogue agents.
\newblock \emph{arXiv preprint arXiv:2503.08026}, 2025.

\bibitem[Tao et~al.()Tao, Xi, Li, Zhao, and Xu]{taohuman}
Z.~Tao, D.~Xi, Z.~Li, J.~Zhao, and W.~Xu.
\newblock Human or llm? a syntactic-semantic collaborative framework for detecting llm-generated peer reviews.
\newblock \emph{A Syntactic-Semantic Collaborative Framework for Detecting Llm-Generated Peer Reviews}.

\bibitem[Wang et~al.(2023{\natexlab{a}})Wang, Li, Chen, Cai, Zhu, Lin, Cao, Liu, Liu, and Sui]{wang2023large}
P.~Wang, L.~Li, L.~Chen, Z.~Cai, D.~Zhu, B.~Lin, Y.~Cao, Q.~Liu, T.~Liu, and Z.~Sui.
\newblock Large language models are not fair evaluators.
\newblock \emph{ArXiv preprint}, abs/2305.17926, 2023{\natexlab{a}}.
\newblock URL \url{https://arxiv.org/abs/2305.17926}.

\bibitem[Wang et~al.(2024{\natexlab{a}})Wang, Tong, Zhang, Li, Zhang, and Chen]{wang2024bpo}
S.~Wang, Y.~Tong, H.~Zhang, D.~Li, X.~Zhang, and T.~Chen.
\newblock Bpo: Towards balanced preference optimization between knowledge breadth and depth in alignment.
\newblock \emph{arXiv preprint arXiv:2411.10914}, 2024{\natexlab{a}}.

\bibitem[Wang et~al.(2023{\natexlab{b}})Wang, Wei, Zhou, Chi, Le, and Schuurmans]{wang2023adversarial}
X.~Wang, J.~Wei, D.~Zhou, E.~Chi, Q.~Le, and D.~Schuurmans.
\newblock Adversarial attacks reveal spurious correlations in large language model evaluations.
\newblock In \emph{Proceedings of the 61st Annual Meeting of the Association for Computational Linguistics (ACL)}, 2023{\natexlab{b}}.

\bibitem[Wang et~al.(2024{\natexlab{b}})Wang, Dong, Delalleau, Zeng, Shen, Egert, Zhang, Sreedhar, and Kuchaiev]{wang2024helpsteer}
Z.~Wang, Y.~Dong, O.~Delalleau, J.~Zeng, G.~Shen, D.~Egert, J.~Zhang, M.~N. Sreedhar, and O.~Kuchaiev.
\newblock Helpsteer 2: Open-source dataset for training top-performing reward models.
\newblock \emph{Advances in Neural Information Processing Systems}, 37:\penalty0 1474--1501, 2024{\natexlab{b}}.

\bibitem[Wang et~al.(2025)Wang, Zeng, Delalleau, Shin, Soares, Bukharin, Evans, Dong, and Kuchaiev]{wang2025helpsteer3}
Z.~Wang, J.~Zeng, O.~Delalleau, H.-C. Shin, F.~Soares, A.~Bukharin, E.~Evans, Y.~Dong, and O.~Kuchaiev.
\newblock Helpsteer3-preference: Open human-annotated preference data across diverse tasks and languages.
\newblock \emph{arXiv preprint arXiv:2505.11475}, 2025.

\bibitem[Wei et~al.()Wei, He, Xia, Liu, Wong, Lin, and Han]{wei2025systematic}
H.~Wei, S.~He, T.~Xia, F.~Liu, A.~Wong, J.~Lin, and M.~Han.
\newblock Systematic evaluation of llm-as-a-judge in llm alignment tasks: Explainable metrics and diverse prompt templates.
\newblock In \emph{ICLR 2025 Workshop on Building Trust in Language Models and Applications}.

\bibitem[Wu et~al.(2024)Wu, Zhan, Wong, Yang, Yang, Yuan, and Chao]{wu2024detectrl}
J.~Wu, R.~Zhan, D.~Wong, S.~Yang, X.~Yang, Y.~Yuan, and L.~Chao.
\newblock Detectrl: Benchmarking llm-generated text detection in real-world scenarios.
\newblock \emph{Advances in Neural Information Processing Systems}, 37:\penalty0 100369--100401, 2024.

\bibitem[Wu and Aji(2025)]{wu2025style}
M.~Wu and A.~F. Aji.
\newblock Style over substance: Evaluation biases for large language models.
\newblock In \emph{Proceedings of the 31st International Conference on Computational Linguistics}, pages 297--312, 2025.

\bibitem[Yang et~al.(2024)Yang, Wu, Ding, Wu, Liang, Gong, Zhang, and Zhang]{yang2025quantifying}
S.~Yang, J.~Wu, W.~Ding, N.~Wu, S.~Liang, M.~Gong, H.~Zhang, and D.~Zhang.
\newblock Quantifying the robustness of retrieval-augmented language models against spurious features in grounding data.
\newblock \emph{arXiv preprint arXiv:2503.05587}, 2024.

\bibitem[Ye et~al.(2024)Ye, Wang, Huang, Chen, Zhang, Moniz, Gao, Geyer, Huang, Chen, et~al.]{ye2024justice}
J.~Ye, Y.~Wang, Y.~Huang, D.~Chen, Q.~Zhang, N.~Moniz, T.~Gao, W.~Geyer, C.~Huang, P.-Y. Chen, et~al.
\newblock Justice or prejudice? quantifying biases in llm-as-a-judge.
\newblock \emph{arXiv preprint arXiv:2410.02736}, 2024.

\bibitem[Yu et~al.()Yu, Luo, Madasu, Lal, and Howard]{yu2024your}
S.~Yu, M.~Luo, A.~Madasu, V.~Lal, and P.~Howard.
\newblock Is your paper being reviewed by an llm? investigating ai text detectability in peer review.
\newblock In \emph{Neurips Safe Generative AI Workshop 2024}.

\bibitem[Yu et~al.(2025)Yu, Luo, Madusu, Lal, and Howard]{yu2025your}
S.~Yu, M.~Luo, A.~Madusu, V.~Lal, and P.~Howard.
\newblock Is your paper being reviewed by an llm? benchmarking ai text detection in peer review.
\newblock \emph{arXiv preprint arXiv:2502.19614}, 2025.

\bibitem[Zellers et~al.(2019)Zellers, Holtzman, Rashkin, Bisk, Farhadi, Roesner, and Choi]{zellers2019grover}
R.~Zellers, A.~Holtzman, H.~Rashkin, Y.~Bisk, A.~Farhadi, F.~Roesner, and Y.~Choi.
\newblock Defending against neural fake news.
\newblock In \emph{Advances in Neural Information Processing Systems}, volume~32, 2019.
\newblock URL \url{https://arxiv.org/abs/1905.12616}.

\bibitem[Zhang et~al.(2024{\natexlab{a}})Zhang, Shang, Wang, Zhang, Yao, Sun, Yu, Yang, and Wei]{zhang2024shifcon}
H.~Zhang, C.~Shang, S.~Wang, D.~Zhang, F.~Yao, R.~Sun, Y.~Yu, Y.~Yang, and F.~Wei.
\newblock Shifcon: Enhancing non-dominant language capabilities with a shift-based contrastive framework.
\newblock \emph{arXiv preprint arXiv:2410.19453}, 2024{\natexlab{a}}.

\bibitem[Zhang et~al.(2024{\natexlab{b}})Zhang, Wu, Li, Yang, Zhao, Jiang, and Tan]{zhang2024balancing}
H.~Zhang, Y.~Wu, D.~Li, S.~Yang, R.~Zhao, Y.~Jiang, and F.~Tan.
\newblock Balancing speciality and versatility: a coarse to fine framework for supervised fine-tuning large language model.
\newblock In \emph{Findings of the Association for Computational Linguistics ACL 2024}, pages 7467--7509, 2024{\natexlab{b}}.

\bibitem[Zhang et~al.()Zhang, Hosseini, Bansal, Kazemi, Kumar, and Agarwal]{zhanggenerative}
L.~Zhang, A.~Hosseini, H.~Bansal, M.~Kazemi, A.~Kumar, and R.~Agarwal.
\newblock Generative verifiers: Reward modeling as next-token prediction.
\newblock In \emph{The Thirteenth International Conference on Learning Representations}.

\bibitem[Zhao et~al.(2025{\natexlab{a}})Zhao, Tan, Ma, Li, Jiang, Wang, Yang, and Liu]{zhao2025chain}
C.~Zhao, Z.~Tan, P.~Ma, D.~Li, B.~Jiang, Y.~Wang, Y.~Yang, and H.~Liu.
\newblock Is chain-of-thought reasoning of llms a mirage? a data distribution lens.
\newblock \emph{arXiv preprint arXiv:2508.01191}, 2025{\natexlab{a}}.

\bibitem[Zhao et~al.(2025{\natexlab{b}})Zhao, Liu, Yu, Kung, Mi, and Yu]{zhao2025one}
Y.~Zhao, H.~Liu, D.~Yu, S.~Kung, H.~Mi, and D.~Yu.
\newblock One token to fool llm-as-a-judge.
\newblock \emph{arXiv preprint arXiv:2507.08794}, 2025{\natexlab{b}}.

\bibitem[Zheng et~al.(2023)Zheng, Chiang, Sheng, Zhuang, Wu, Zhuang, Lin, Li, Li, Xing, et~al.]{zheng2023judging}
L.~Zheng, W.-L. Chiang, Y.~Sheng, S.~Zhuang, Z.~Wu, Y.~Zhuang, Z.~Lin, Z.~Li, D.~Li, E.~Xing, et~al.
\newblock Judging llm-as-a-judge with mt-bench and chatbot arena.
\newblock \emph{Advances in Neural Information Processing Systems}, 36:\penalty0 46595--46623, 2023.

\bibitem[Zhu et~al.(2025)Zhu, Weng, Yang, and Zhang]{zhu2025deepreview}
M.~Zhu, Y.~Weng, L.~Yang, and Y.~Zhang.
\newblock Deepreview: Improving llm-based paper review with human-like deep thinking process.
\newblock \emph{arXiv preprint arXiv:2503.08569}, 2025.

\end{thebibliography}
\appendix

\section{The Use of LLMs for Writing}
We employed Google's Gemini 2.5 Pro and OpenAI's GPT-5 as writing assistance tools during the preparation of this manuscript. Their role was exclusively for language refinement, such as improving readability and rephrasing for clarity in an academic writing style. This usage aligns with standard academic practices for language polishing.

\section{Experiment Implementation Details}
\label{Experiment Implementation Details}

\subsection{Detailed Definition of Various Judgment Types}
Depending on the evaluation protocol, judgments can take multiple forms~\citep{li2024generation}: (i) \textit{Score-based judgments}: $j \in \mathbb{R}$, such as a numerical rating on one or several dimensions; (ii) \textit{Pairwise judgments}: $j \in \{(c_a \succ c_b), (c_b \succ c_a)\}$, indicating a preference between two candidates $c_a, c_b \in \mathcal{C}$; (iii) \textit{Listwise judgments}: $j \in \pi(\mathcal{C})$, representing a permutation (ranking) $\pi$ over a candidate set.

\subsection{\textit{JD-Bench Details}}

To systematically study the detectability of LLM-generated judgments, we introduce \textbf{JD-Bench}, a large-scale benchmark that integrates diverse applications, judgment types, and model sources. JD-Bench provides a unified testbed for evaluating both existing and newly proposed detectors under realistic settings.

\paragraph{Dataset Selection.}
We construct JD-Bench by aggregating data from multiple domains and judgment types, ensuring broad coverage of evaluation practices:
\begin{itemize}[leftmargin=0pt, itemsep=0pt, topsep=0pt, parsep=0pt, partopsep=0pt]
\item \textbf{HelpSteer2} \citep{wang2024helpsteer}: HelpSteer2 is an open-source dataset designed to train and evaluate reward models for helpfulness assessment of LLM-generated responses. It contains large-scale human-annotated pointwise judgments that assign numerical scores to responses across diverse instruction-following tasks. The dataset covers multiple domains and languages, enabling robust generalization of reward models. Its fine-grained annotations make it a strong benchmark for pointwise/score-based evaluation.
\item \textbf{HelpSteer3} \citep{wang2025helpsteer3}: HelpSteer3 extends HelpSteer2 by collecting pairwise human preference data on LLM responses. Instead of absolute scores, annotators compare two candidate responses to the same prompt and indicate which is better, yielding high-quality comparative judgments. The dataset spans a wide range of tasks and languages, supporting cross-lingual preference modeling and fine-grained ranking evaluation.

\item \textbf{NeurIPS Review Dataset} \citep{yu2025your}: This dataset comprises a large collection of real academic peer reviews from the NeurIPS conference, annotated with multi-dimensional scores such as soundness, novelty, clarity, and overall rating. It represents a domain where judgments are structured, multi-faceted, and highly consequential. The dataset captures nuanced reviewing language and decision rationales, providing a challenging benchmark for modeling human-like expert evaluation. It is especially valuable for studying judgment behavior in formal and high-stakes settings.

\item \textbf{ANTIQUE} \citep{hashemi2020antique}: ANTIQUE is a benchmark for non-factoid question answering, focused on ranking passages based on their relevance to user queries. It includes listwise relevance judgments collected from crowdworkers, where multiple candidate documents are ordered according to their usefulness. The questions are open-ended and require deeper understanding rather than simple fact retrieval, making the ranking task more challenging.

\end{itemize}

Each dataset provides \emph{human-labeled} judgments as a reliable reference. To complement these, we collect \emph{LLM-generated} judgments following the judging principles outlined in the respective papers, ensuring consistency in evaluation criteria.

\paragraph{LLM Selection.} 
To obtain LLM-generated judgments, we employ a diverse set of both closed-source and open-source models across a wide range of sizes and model families. This diversity is essential to cover heterogeneous judgment patterns and to test detector generalization. Specifically, JD-Bench includes judgments from:
\begin{itemize}
    \item \textbf{Closed-source models:}
    \begin{itemize}
        \item OpenAI series: GPT-4o, GPT-5-mini, o4-mini.
        \item Anthropics series: Claude-Haiku-3.5, Claude-Haiku-3, Claude-3.5-Sonnet.
        \item Google series: Gemini-2.0-Flash, Gemini-2.5-Flash, Gemini-1.5-Pro.
    \end{itemize}

    \item \textbf{Open-source models:} 
    \begin{itemize}
        \item LLaMA family: LLaMA-3.2-3B, LLaMA-3.1-8B, LLaMA-3.1-70B.
        \item Qwen family: Qwen-2.5-7B, Qwen-2.5-72B, RM-R1, RISE-Judge.
        \item Mistral family: Mistral-7B, Mistral-Small-24B.
        \item DeepSeek series: DeepSeek-V3, DeepSeek-R1.
    \end{itemize}
\end{itemize}

This mixture of datasets and models results in a benchmark that is both large-scale and diverse: JD-Bench covers \emph{multiple application scenarios}, \emph{different judgment types} (score, pairwise, listwise), and \emph{a wide spectrum of LLM families}, making it a comprehensive resource for advancing judgment detection research. Table~\ref{tab:jd-bench-datasets} presents the statistics of JD-Bench.

\begin{table}[t]
\small
\centering
\setlength{\tabcolsep}{4pt}
\caption{Overview of datasets included in JD-Bench.}
\renewcommand{\arraystretch}{1.25}
\begin{tabular}{l p{2.4cm} p{2.4cm} p{2.4cm} p{2.4cm}}
\toprule
\textbf{Dataset} & \textbf{HelpSteer2} & \textbf{HelpSteer3} & \textbf{NeurIPS} & \textbf{ANTIQUE} \\
\midrule
Application & Resp. Eval. & Resp. Eval. & Peer Review & Doc Ranking \\
Judgment Type & Pointwise & Pairwise & Pointwise & Listwise \\
Judgment Dims & Helpfulness, Correctness, Coherence, Complexity, Verbosity 
& Overall 
& Overall, Confidence, Soundness, Presentation, Contribution 
& Relevance \\
Rating Scale & 0–4 & –3–3 & 1–10 / 1–5 / 1–4 & 1–4 \\
\#Train / \#Test & 62,961 / 21,778 & 62,880 / 42,317 & 63,210 / 62,664 & 102,417 / 61,909 \\
\bottomrule
\end{tabular}
\label{tab:jd-bench-datasets}
\end{table}

\paragraph{Prompt for JD-Bench Construction}

\begin{tcolorbox}[breakable, title={HelpSteer2 Prompt (Pointwise, 5-Dimension Scoring)}, label=hs2:pointwise]
\small
\ttfamily
Given a prompt and a response, follow the rubric to make a judgment.

\vspace{3mm}
\#\# Rubric:

Judge the response on five aspects: \textbf{helpfulness}, \textbf{correctness}, \textbf{coherence}, \textbf{complexity}, and \textbf{verbosity}.\

Assign each aspect a scalar score in \textbf{[0, 4]}.

\vspace{3mm}
\#\# Prompt: [PROMPT]

\vspace{3mm}
\#\# Response: [RESPONSE]

\vspace{3mm}
Please output a valid JSON object using the following schema:
{
\hspace{3mm}"Rationale": \textless explanation for the given scores\textgreater,\
\hspace{3mm}"Helpfulness": \textless 0--4\textgreater,\
\hspace{3mm}"Correctness": \textless 0--4\textgreater,\
\hspace{3mm}"Coherence": \textless 0--4\textgreater,\
\hspace{3mm}"Complexity": \textless 0--4\textgreater,\
\hspace{3mm}"Verbosity": \textless 0--4\textgreater
}

\vspace{3mm}
Formatted the abovementioned schema and produce the judgment JSON now.
\end{tcolorbox}

\begin{tcolorbox}[breakable, title={HelpSteer3 Prompt (Pairwise Comparison)}, label=hs3:pairwise]
\small
\ttfamily
Given a prompt and two responses, follow the rubric to make a comparative judgment.

\vspace{3mm}
\#\# Rubric:
Compare \textbf{Response 1} and \textbf{Response 2} along five aspects: \textbf{helpfulness}, \textbf{correctness}, \textbf{coherence}, \textbf{complexity}, and \textbf{verbosity}.\
Assign a single comparative score in {-3,-2,-1,0,1,2,3} using the scale: \
-3: R1 \textit{much better} than R2; \ \ -2: R1 \textit{better} than R2; \ \ -1: R1 \textit{slightly better} than R2; \
\phantom{-3:}\ 0: about the same; \ \ 1: R2 \textit{slightly better} than R1; \ \ 2: R2 \textit{better} than R1; \ \ 3: R2 \textit{much better} than R1.

\vspace{3mm}
\#\# Prompt (conversation/context):\
[CONTEXT AS FLATTENED TEXT]

\vspace{3mm}
\#\# Response 1:\
[RESPONSE\_1]

\vspace{3mm}
\#\# Response 2:\
[RESPONSE\_2]

\vspace{3mm}
Please output a valid JSON object using the following schema:
{
\hspace{3mm}"Rationale": \textless explanation for the comparative score\textgreater,\
\hspace{3mm}"Score": \textless -3\textbar -2\textbar -1\textbar 0\textbar 1\textbar 2\textbar 3\textgreater
}

\vspace{3mm}
Formatted the abovementioned schema and produce the judgment JSON now.
\end{tcolorbox}
\begin{tcolorbox}[breakable, title={NeurIPS Review Prompt (Structured JSON Review)}, label=neurips:review]
\small
\ttfamily
You are an AI researcher reviewing a paper submitted to a prestigious AI conference. Thoroughly evaluate the paper, adhering to the provided guidelines, and return a detailed assessment in the specified JSON format.

\vspace{3mm}
\#\# Manuscript: [MANUSCRIPT TEXT OR CONCATENATED CHUNKS]

\vspace{3mm}
\#\# Reviewer Guidelines (dimensions to cover):

\textbf{Summary}: Briefly summarize contributions (no critique here).\

\textbf{Strengths \& Weaknesses} across: Originality, Quality, Clarity, Significance.\

Provide \textbf{Questions} for authors (useful for rebuttal).\

Discuss \textbf{Limitations} and potential societal impact.\

Flag \textbf{Ethical Concerns} if applicable (per conference policy).\

Assign numerical ratings: \textbf{Soundness}, \textbf{Presentation}, \textbf{Contribution} (1--4 each).\

Provide an \textbf{Overall} score (1--10) and \textbf{Confidence} (1--5).

\vspace{3mm}
\#\# Output a valid JSON object with the following fields:
{
\hspace{3mm}"Summary": \textless summary for the paper\textgreater,\
\hspace{3mm}"Questions": \textless questions for the author\textgreater,\
\hspace{3mm}"Limitations": \textless limitations for the paper\textgreater,\
\hspace{3mm}"Soundness": \textless 1--4\textgreater,\
\hspace{3mm}"Presentation": \textless 1--4\textgreater,\
\hspace{3mm}"Contribution": \textless 1--4\textgreater,\
\hspace{3mm}"Overall": \textless 1--10\textgreater,\
\hspace{3mm}"Confidence": \textless 1--5\textgreater
}

\vspace{3mm}
Formatted the abovementioned schema and produce the review JSON now.
\end{tcolorbox}

\begin{tcolorbox}[breakable, title={ANTIQUE Prompt (3-Way Relevance Ranking)}, label=antique:rank3]
\small
\ttfamily
Given a prompt and three responses, follow the rubric to assess relevance and rank the responses.

\vspace{3mm}
\#\# Rubric (per-response relevance score in [1, 4]):\
\textbf{4}: Reasonable and convincing; on par with or better than a likely correct answer.\
\textbf{3}: Possibly an answer, but not sufficiently convincing; a better-quality answer likely exists.\
\textbf{2}: Not an acceptable answer; unreasonable or does not address the question, but still on-topic.\
\textbf{1}: Completely out of context or nonsensical.

\vspace{3mm}
\#\# Prompt: [QUERY]

\vspace{3mm}
\#\# Response 1: [RESPONSE\_1]

\vspace{3mm}
\#\# Response 2: [RESPONSE\_2]

\vspace{3mm}
\#\# Response 3: [RESPONSE\_3]

\vspace{3mm}
Please output a valid JSON object using the following schema:
{
\hspace{3mm}"Rationale": \textless explanation for your judgment and ranking\textgreater,\
\hspace{3mm}"Response1 Score": \textless 1--4\textgreater,\
\hspace{3mm}"Response2 Score": \textless 1--4\textgreater,\
\hspace{3mm}"Response3 Score": \textless 1--4\textgreater,\
\hspace{3mm}"Ranking": \textless list of indices indicating best$\rightarrow$worst, e.g., [0,1,2]\textgreater
}

\vspace{3mm}
Formatted the abovementioned schema and produce the judgment JSON now.
\end{tcolorbox}

\subsection{J-Detector Details}

\subsubsection{Linguistic Features}

We extract a comprehensive set of surface, lexical, syntactic, and discourse indicators from each candidate response using \texttt{spaCy}-based parsing pipelines.

\begin{itemize}[leftmargin=1.5em]
    \item \textbf{Length \& Structure:}
    \texttt{word\_count}, \texttt{char\_count}, \texttt{sentence\_count}, \texttt{avg\_sentence\_length},
    \texttt{list\_count} (bullet or numbered lists), \texttt{paragraph\_count}, \texttt{punctuation\_count}, \texttt{reference\_count} (e.g., URLs).
    \item \textbf{Lexical Diversity:}
    \texttt{unique\_words}, \texttt{vocab\_diversity} (unique/total word ratio), \texttt{average\_word\_length}, \texttt{noun\_verb\_ratio}, \texttt{adjective\_ratio}, \texttt{adverb\_ratio}, \texttt{pronoun\_ratio}, \texttt{contraction\_rate}.
    \item \textbf{Readability:}
    \texttt{coleman\_liau} index.
    \item \textbf{Syntactic Complexity:}
    \texttt{syntax\_tree\_depth} (maximum dependency depth), \texttt{average\_dependency\_length}, \texttt{passive\_voice\_ratio} (fraction of sentences with \texttt{nsubjpass}/\texttt{csubjpass}), \texttt{subordinate\_clause\_rate} (rate of \texttt{mark} tokens).
    \item \textbf{Discourse/hedging:}
    \texttt{hedging\_frequency} (occurrence of hedge words such as ``may'', ``possibly''), \texttt{discourse\_marker\_rate} (connectives such as ``however'', ``moreover'').
\end{itemize}

These features are computed for each response independently. For pairwise or listwise datasets (e.g., HelpSteer3, ANTIQUE), we additionally compute \emph{difference features} such as \(r_1 - r_2\) on each scalar dimension when comparing two responses.


\subsubsection{LLM-Enhanced Features}

Beyond surface-level indicators, we harness powerful large language models (e.g., Qwen3-8B) to derive task-aligned evaluation features. For each dataset, the model is prompted with the original instruction or query together with its candidate responses, and asked to generate structured JSON judgments that include detailed rationales and aspect-specific scores.

\paragraph{Pointwise Setting (e.g., HelpSteer2).}  
Each response is scored independently along eight stylistic and content dimensions:
\begin{itemize}[leftmargin=1.5em]
    \item Style, Format, Wording
    \item Helpfulness, Correctness, Coherence
    \item Complexity, Verbosity
\end{itemize}
The model outputs both a natural language rationale and numeric scores (0–4) per dimension plus an \texttt{overall\_score}.

\paragraph{Pairwise Setting (e.g., HelpSteer3).}  
Two responses are jointly compared under criteria such as \emph{helpfulness}, \emph{correctness}, \emph{coherence}, \emph{complexity}, and \emph{verbosity}. The LLM produces a signed comparison score from \(-3\) (Response~1 $\gg$ Response~2) to \(+3\) (Response~2 $\gg$ Response~1) and a supporting rationale.

\paragraph{Listwise Setting (e.g., ANTIQUE).}  
Three responses are simultaneously ranked by relevance. The LLM assigns a 1–4 relevance score to each response and outputs an ordered ranking list \([0,1,2]\) to indicate relative quality.

\paragraph{Long-form Paper Evaluation (e.g., NeurIPS Submissions).}  
For full papers, we ask the model to return review-like signals: style, format, wording (0–4), rating (1–10), confidence (1–5), soundness/presentation/contribution (1–4 each), together with detailed reasoning.

\begin{table}[h!]
\centering
\scriptsize
\caption{Example LLM-enhanced feature dimensions by dataset.}
\begin{tabular}{ll}
\toprule
\textbf{Dataset Setting} & \textbf{LLM-Generated Feature Dimensions} \\
\midrule
HelpSteer2 (pointwise) & Style, Format, Wording, Helpfulness, Correctness, Coherence, Complexity, Verbosity, Overall \\
HelpSteer3 (pairwise) & Helpfulness, Correctness, Coherence, Complexity, Verbosity, Pairwise Score (\(-3\)~--~\(+3\)) \\
ANTIQUE (listwise) & Response relevance scores (1--4), Ranking order, Rationale \\
NeurIPS (pointwise) & Style, Format, Wording, Rating (1--10), Confidence (1--5), Soundness, Presentation, Contribution \\
\bottomrule
\end{tabular}
\end{table}

These LLM-enhanced features provide semantically rich, high-level signals that complement the surface-level linguistic statistics, enabling our detector to exploit both human-interpretable cues and task-specific, model-derived evaluations.

\subsection{SLM-based Method Details}

To benchmark the ability of small language models (SLMs) to discriminate between human and LLM-generated judgments, we adapt text classification pipelines with two input configurations: \emph{judgment-only} (w/o candidates) and \emph{judgment+candidate} (w/ candidates). Both settings train a binary classifier to predict whether a group of judgments originates from a human annotator (label $0$) or an LLM (label $1$). We employ \texttt{roberta-base} and \texttt{allenai/longformer-base-4096} as backbones, with max sequence lengths $512$ and $4096$, respectively.

\begin{itemize}
    \item \textbf{Judgment-Only}  
    Inspired by SLM-based text detection, this setting feeds only the \emph{judgment artifacts} into the model. Each group is represented by a textualized summary of available signals, including:
    \begin{itemize}
        \item \emph{Numeric scores}: fields such as \texttt{rating}, \texttt{score}, \texttt{confidence}, \texttt{soundness}, \texttt{presentation}, \texttt{contribution}, etc.
        \item \emph{Pairwise comparisons}: keys such as \texttt{pairwise}, \texttt{pairs}, \texttt{comparisons}, or \texttt{prefs}.
        \item \emph{Ranking lists}: an explicit \texttt{ranking} field if available.
        \item \emph{Metadata}: optional question/prompt/task descriptions to provide minimal context.
    \end{itemize}
    The resulting text is tokenized and directly used as the classifier input.
    
    \item \textbf{Judgment + Candidate}  
    In this richer setting, we augment the above judgment text with the \emph{candidate contents} being judged. Candidate responses are extracted from dataset fields such as:
    \begin{itemize}
        \item \texttt{examples[*].docs} for passage-style corpora (e.g., ANTIQUE);
        \item \texttt{examples[*].context} for conversational datasets (e.g., HelpSteer3), where only assistant turns are kept;
        \item top-level \texttt{docs}, \texttt{candidates}, or \texttt{answers} if present.
    \end{itemize}
    Since candidate texts can be long, we apply a \emph{head+tail trimming} strategy per candidate to respect the model's maximum input length. Judgment tokens are prioritized to remain intact. The final input is a concatenation:
    \[
        \texttt{JudgmentText} \;||\; \texttt{=== Candidates ===} \;||\; \texttt{Candidate}_1 \;||\; \dots \;||\; \texttt{Candidate}_n.
    \]
\end{itemize}

\begin{table}[h]
\centering
\small
\begin{tabular}{lll}
\toprule
\textbf{Mode} & \textbf{Input Composition} & \textbf{Example Fields Used} \\
\midrule
w/o candidates & Judgments only & ratings, scores, pairwise, ranking, task \\
w/ candidates  & Judgments + trimmed candidate texts & docs, context (assistant turns), answers \\
\bottomrule
\end{tabular}
\caption{Two input modes for SLM-based judgment detection.}
\label{tab:detector_modes}
\end{table}

During training, both settings use the HuggingFace \texttt{Trainer} with standard hyperparameters (\texttt{AdamW}, learning rate $2\times 10^{-5}$, batch size $8$, weight decay $0.01$). Labels are mapped to $\{0,1\}$, with \texttt{Human}$\mapsto0$ and \texttt{LLM}$\mapsto1$. Evaluation reports accuracy, F1, and AUROC on held-out test splits.

\subsection{LLM-based Method Details}

\subsubsection{LLM-as-a-Judge Detector}
\label{sup:llm-judge-detector}

Inspired by logits-based AI-generated text detection~\citep{mitchell2023detectgpt}, we design a \textbf{single-pass detector} that treats an LLM as a surrogate judge.  
Given a group of judgments $G$, we build a compact textual payload including:
\begin{itemize}[leftmargin=1.5em]
    \item \textbf{Judgment-only signals:} helpfulness, correctness, coherence, complexity, verbosity, ranking, and pairwise preferences.
    \item \textbf{Optional candidates:} trimmed prompt/response or passage text to provide weak context.
\end{itemize}

We prompt the detector LLM with an instruction template asking it to decide whether the judgments were written by a \textit{Human} or by an \textit{LLM}, based on style, consistency, and calibration artifacts:
\begin{verbatim}
{
  "Rationale": "<brief explanation>",
  "Prediction": "Human" | "LLM"
}
\end{verbatim}

Two modes are supported:
\begin{itemize}[leftmargin=1.5em]
    \item \texttt{judgment\_only}: only judgment artifacts are provided.
    \item \texttt{enable\_candidate}: judgment artifacts plus trimmed candidate texts.
\end{itemize}

This baseline does not use any explicit feature engineering but leverages the LLM's implicit ability to reason about stylistic and distributional cues.

\subsubsection{Sample-Level LLM-Based Analysis}
\label{sup:sample-analysis}

We further design an \textbf{agentic feature mining} procedure to expose regularities in Human vs.\ LLM judgments at the \emph{instance level}.  
Given a training set of groups, we:
\begin{enumerate}[leftmargin=1.5em]
    \item Flatten them into a table of \emph{prompt, response, label, scores}, and derived metrics such as length and average score.
    \item Mine \textbf{Human–LLM pairs} using two strategies:
    \begin{itemize}[leftmargin=1.2em]
        \item \texttt{scoring}: select $k$ pairs with the largest average-score gaps under the same prompt.
        \item \texttt{pairwise}: sample $k$ random Human–LLM pairs.
    \end{itemize}
    \item Feed each pair to an LLM agent that proposes actions to maintain a \textbf{Feature Bank}:
    \begin{verbatim}
    Add:    {"name": "...", "description": "..."}
    Delete: {"name": "..."}
    Merge:  {"name": "...", "description": "...", "existing": "..."}
    \end{verbatim}
    \item Typical mined features include:
    \begin{itemize}[leftmargin=1.2em]
        \item Length or verbosity bias;
        \item Overly smooth or formulaic score patterns;
        \item Deterministic tone and calibration artifacts.
    \end{itemize}
\end{enumerate}

The resulting \textbf{Feature Bank} $\mathcal{F}_{\text{sample}}$ captures diagnostic cues distilled by the LLM itself and is later injected into the final detection prompt.

\subsubsection{Distribution-Level LLM-Based Analysis}
\label{sup:distribution-analysis}

Beyond individual samples, we analyze \textbf{dataset-wide statistics} to extract global signals of LLM-generated judgments:
\begin{enumerate}[leftmargin=1.5em]
    \item Compute per-label histograms and descriptive statistics for all available judgment dimensions (e.g., helpfulness, correctness, coherence, complexity).
    \item Analyze \textbf{correlations}:
    \begin{itemize}[leftmargin=1.2em]
        \item Length–score Spearman correlations within Human/LLM groups;
        \item Cross-dimension correlations (e.g., helpfulness vs.\ coherence).
    \end{itemize}
    \item Summarize these findings as structured text and feed them to an LLM to propose additional high-level features, such as:
    \begin{itemize}[leftmargin=1.2em]
        \item Consistent score calibration (LLM often shows smaller variance);
        \item Stronger length–score coupling in LLM judgments;
        \item Reduced inter-dimension diversity compared to human raters.
    \end{itemize}
\end{enumerate}

The discovered global patterns augment the feature bank as $\mathcal{F}_{\text{dist}}$, complementing sample-level cues with distributional regularities.

\subsubsection{Final Detection}
\label{sup:final-detection}

The final detector integrates:
\begin{itemize}[leftmargin=1.5em]
    \item A \textbf{Feature Bank} $\mathcal{F}=\mathcal{F}_{\text{sample}}\cup\mathcal{F}_{\text{dist}}$;
    \item Group-level summaries (judgments + optional candidates).
\end{itemize}
An LLM receives this structured prompt and outputs the final label prediction:
\[
\hat{y} = f_{\text{LLM}}(\mathrm{summary}(G), \mathcal{F}),
\]
where $f_{\text{LLM}}$ denotes the LLM-based reasoning process conditioned on both the mined features and the group payload.

\begin{table}[h!]
\centering
\small
\caption{Comparison of the three LLM-based detection strategies.}
\begin{tabular}{lccc}
\toprule
\textbf{Method} & \textbf{Uses Candidates?} & \textbf{Feature Bank} & \textbf{Level of Analysis} \\
\midrule
LLM-as-a-Judge & Optional & None & Per-group \\
Sample-level   & Optional & $\mathcal{F}_{\text{sample}}$ & Instance-level \\
Distribution-level & Optional & $\mathcal{F}_{\text{sample}}+\mathcal{F}_{\text{dist}}$ & Global + per-group \\
\bottomrule
\end{tabular}
\end{table}

In practice, the multilevel detector (sample + distribution) consistently improves accuracy by guiding the LLM with both fine-grained instance cues and global dataset regularities.

\section{Theoretically Analysis on LLM-generated Judgment Detectability}
\label{Theoretically Analysis on LLM-generated Judgment Detectability}

We model the detectability of whether a group of judgments $G$ (scores, pairwise preferences, or listwise rankings) was produced by a human or an LLM.  
Let $m$ denote the group size, $d$ the number of attribute dimensions, and $S$ the effective rating scale cardinality:
\[
S =
\begin{cases}
L, & \text{for $L$-level scoring;}\\[2pt]
2x+1, & \text{for pairwise judgments with $x \in \mathbb{Z}_{\ge 1}$ superiority levels per side (including tie);}\\[2pt]
k!, & \text{for a full ranking over $k$ candidates.}
\end{cases}
\]
The per-judgment information is $\log S$ nats.\footnote{For listwise $k!$, Stirling’s approximation gives $\log(k!) \approx k\log k - k$. For continuous pairwise margins, discretization into $B$ bins yields $S = B$.}

Let $P_H$ and $P_M$ be the conditional distributions over judgment outcomes induced by humans and LLMs, respectively.  
Denote $\Delta = \mathrm{TV}(P_H,P_M)$ as their total variation distance.  

\paragraph{From sample complexity to group detectability.}  
With $n$ i.i.d.\ observations, the total variation between product distributions grows as
\[
\mathrm{TV}\big(P_H^{\otimes n}, P_M^{\otimes n}\big)
= 1 - \exp\{-n I_c(P_H,P_M) + o(n)\},
\]
where $I_c$ is the Chernoff information, scaling quadratically with $\Delta$.  
In our setting, the effective observation budget is
\[
n_{\text{eff}} = m \cdot d \cdot \log S,
\]
which accounts for group size, dimensionality, and rating resolution.

\paragraph{Detectability index.}  
Thus, the detectability index becomes
\[
\mathsf{Det}(G)
= 1 - \exp\{-\,\beta \, m d \log S \, \Delta^2\},
\]
where $\beta>0$ is dataset- and model-dependent.  
The detectability increases monotonically with four factors:
(i) rating scale $S$, (ii) attribute dimensions $d$, (iii) group size $m$, and (iv) distribution gap $\Delta$.

\paragraph{Instantiation by type.}  
For $L$-level scores, use $S=L$.  
For pairwise preferences, use $S=L_{\text{pair}}$ (e.g., $7$ for \{-3,\dots,3\}).  
For listwise ranking over $k$ items, use $S=k!$ (or $\log S \approx k\log k - k$).  
For mixed-type groups, sum $m d \log S$ across instances. 

\end{document}